\documentclass[letterpaper]{article}

\newcommand{\comment}[1]{}

\usepackage[final]{neurips_2022}

\usepackage[utf8]{inputenc} 
\usepackage[T1]{fontenc}    
\usepackage[hidelinks]{hyperref}      
\usepackage{url}            
\usepackage{booktabs}       
\usepackage{amsfonts}       
\usepackage{nicefrac}       
\usepackage{microtype}      
\usepackage{xcolor}         
\usepackage{graphicx}
\usepackage{bbding} 
\usepackage{array}
\usepackage{amsmath}

\newcommand\blfootnote[1]{%
  \begingroup
  \renewcommand\thefootnote{}\footnote{#1}%
  \addtocounter{footnote}{-1}%
  \endgroup
}

\usepackage[shortlabels]{enumitem}
\usepackage{placeins}
\usepackage{dirtree}
\usepackage{multirow}
\usepackage{biblatex}
\bibliography{citations.bib}

\newcommand{\hiddensubsection}[1]{
    \refstepcounter{subsection}
    \subsection*{\arabic{subsection}\hspace{1em}{#1}}
}
\newcommand{\hiddensection}[1]{
    \refstepcounter{section}
    \section*{\arabic{section}\hspace{1em}{#1}}
}

\title{PulseImpute: A Novel Benchmark Task for \\
Pulsative Physiological Signal Imputation}

\author{
  Maxwell A. Xu\textsuperscript{1},
  Alexander Moreno\textsuperscript{1*, 2},
  Supriya Nagesh\textsuperscript{1},
  V. Burak Aydemir\textsuperscript{1}, \and
  \textbf{David W. Wetter\textsuperscript{3},
  Santosh Kumar\textsuperscript{4},
  James M. Rehg\textsuperscript{1}}  \and
  \textsuperscript{1 }Georgia Tech, \textsuperscript{2 }Luminous Computing, \textsuperscript{3 }University of Utah, \textsuperscript{4 }University of Memphis \and
  \texttt{\{maxxu,...,rehg\}@gatech.edu}
}

\begin{document}

\maketitle

\begin{abstract}
The promise of Mobile Health (mHealth) is the ability to use wearable sensors to monitor participant physiology at high frequencies during daily life to enable temporally-precise health interventions. However, a major challenge is frequent missing data. Despite a rich imputation literature, existing techniques are ineffective for the pulsative signals which comprise many mHealth applications, and a lack of available datasets has stymied progress.
We address this gap with \emph{PulseImpute}, the first large-scale pulsative signal imputation challenge which includes realistic mHealth missingness models, an extensive set of baselines, and clinically-relevant downstream tasks. Our baseline models include a novel transformer-based architecture designed to exploit the structure of pulsative signals.
We hope that PulseImpute will enable the ML community to tackle this significant and challenging task.
\end{abstract}
\hiddensection{Introduction}

The goal of mobile health (mHealth) is to use continuously collected signals from wearable devices, such as smart watches, to passively monitor a user's health states during daily life and deliver interventions to improve health outcomes. The use of devices such as Fitbit to monitor physical activity has now become an established practice, with large-scale consumer adoption. Even more exciting is the increasing feasibility of measuring complex health states, such as stress \cite{cstress}, by leveraging high-frequency physiological signals from wearable sensing technologies such as photoplethysmography (PPG) or electrocardiography (ECG). A subset of these physiological signals are \emph{pulsative}, which we define as signals that have a quasiperiodic structure with specific signal morphologies (e.g. the QRS complex in ECG), which vary over time and across populations due to their origins in the cardiopulmonary system. The rich structure of these signals in terms of shape and timing has significant clinical value, for tasks such as heart disease diagnosis \cite{vanderbilt2022myo}. \blfootnote{$^*$Work was done while the author was at Georgia Tech.}

However, a key challenge is addressing \emph{missing data}, which is commonplace and arises from multiple causes such as insecure sensor attachment or data transmission loss~\cite{Rahman2017mdebugger}. Current mHealth systems 
either employ simple imputation methods, such as KNN \cite{activity_knn}, 
or simply do not trigger health interventions when inputs are missing \cite{cstress_knn}. 
Since mHealth biomarkers may require multiple signals as input \cite{cstress}, the latter approach can lead to long intervals of missingness due to the juxtaposition of missingness patterns across the inputs.
However, the quasiperiodic nature of these signals provides rich information for imputation, 
which can be exploited by modeling morphological structures over time. Additionally, the accuracy with which a signal's morphology can be recovered has a direct impact on downstream task performance, as specific morphological properties have clinical significance.
Furthermore, the well-defined signal morphologies make it easy to interpret reconstruction results. 
Thus, it follows that pulsative signals provide a novel context for the development of ML imputation methods, especially in comparison to prior imputation tasks. 

We introduce \emph{PulseImpute}, a novel pulsative signal imputation challenge to catalyze and enable the ML community to address the important missing data problems underlying current and future mHealth applications. Table~\ref{tbl:necessarycomp} describes six criteria that the PulseImpute challenge provides, which no prior works address in full. 
We extract real missingness patterns from real-world mHealth field studies \cite{chatterjee2020smokingopp, reiss2019ppgdalia} and mimic specific mHealth missingness paradigms \cite{Rahman2017mdebugger} in order to apply these patterns to open source pulsative signal datasets. As a result, we can simulate realistic missingness while using the original ablated samples as ground truth, making it possible to quantify and visualize the accuracy of imputation. We also include three downstream tasks, Heartbeat Detection in ECG/PPG and Cardiac Pathophysiology Multi-label Classification in ECG, making it possible to quantify the impact of imputed values on downstream performance. PulseImpute also features an extensive benchmark suite of imputation methods covering both traditional and deep-learning-based approaches. In particular, we introduce a novel transformer imputation baseline with a Bottleneck Dilated Convolution (BDC) self-attention module that is designed for the pulsative signal structure and provides state-of-the-art (SOTA) performance. These baselines provide a strong context for future research efforts.

To summarize, we make the contribution of introducing the PulseImpute Challenge, which is composed of 1) a comprehensive benchmark suite for mHealth pulsative signal imputation with publicly-available data across multiple signal modalities and reproducible missingness models; 2) nine baseline models which demonstrate the failure of existing time-series imputation methods to address our novel challenge; and 3) an additional novel baseline incorporating a self-attention module which learns to attend to quasiperiodic features and delivers SOTA performance. The benchmarking code and datasets can be found at \url{www.github.com/rehg-lab/pulseimpute} and \url{www.doi.org/10.5281/zenodo.7129965}, respectively.

\renewcommand{\arraystretch}{1.2}
\begin{table}[]
\begin{center}
\resizebox{.8\textwidth}{!}{%
\begin{tabular}{|*{6}{>{\centering\arraybackslash}m{.2\linewidth}|}}
\hline
 & \begin{tabular}[c]{@{}c@{}}\textit{\textbf{Our PulseImpute}}\\ \textit{\textbf{Challenge}}\end{tabular} & \begin{tabular}[c]{@{}c@{}}\textit{Standard Imputation} \\ \textit{Datasets} \cite{physionet2012, kdd}  \comment{\cite{physionet2012, kdd}} \end{tabular} & \begin{tabular}[c]{@{}c@{}}\textit{mHealth Systems'}\\ \textit{Imputation} \cite{iranfar2021relearn, cstress_knn}\end{tabular} & \begin{tabular}[c]{@{}c@{}}\textit{Pulsative Signal} \\ \textit{Imputation} \cite{rpca, varim}  \end{tabular} \\ \hline
mHealth Pulsative Signals & \LARGE\checkmark &   & \LARGE\checkmark &  \\ \hline 
Publicly Available Data & \LARGE\checkmark  & \LARGE\checkmark &  & \LARGE\checkmark \\ \hline
Realistic Missingness & \LARGE\checkmark & \LARGE\checkmark & \LARGE\checkmark &  \\ \hline
Directly Evaluates Imputation & \LARGE\checkmark  & \LARGE\checkmark &  & \LARGE\checkmark \\ \hline
Comprehensive Benchmarks 
& \LARGE\checkmark & \LARGE\checkmark &  &  \\ \hline
Downstream  \ \ \ \ \ \ \ \ Tasks & \LARGE\checkmark & \LARGE\checkmark & \LARGE\checkmark & \LARGE\checkmark \\ \hline
\end{tabular}%
} \vspace{.1cm}
\caption{Necessary components for an mHealth pulsative signal imputation challenge.  
Our PulseImpute Challenge is the only work to meet all six criteria.}
\label{tbl:necessarycomp}
\end{center}
\vspace{-.8cm}
\end{table}

\begin{figure*}[]
\begin{center}
\includegraphics[width=\textwidth]{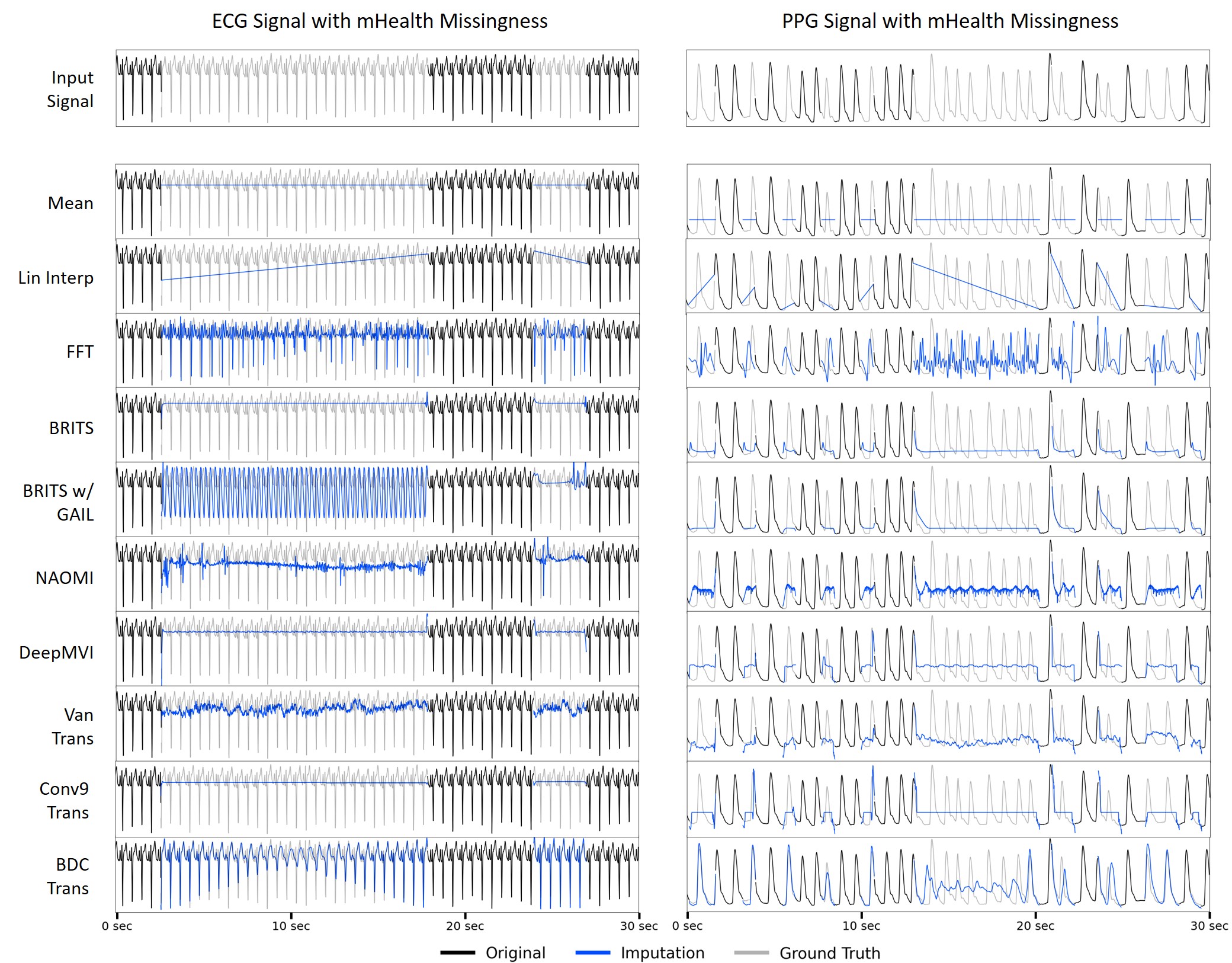} 
\caption{Visualization of imputation results on ECG and PPG signals. The large gaps found in real-world missingness patterns create substantial challenges for all methods. BRITS and DeepMVI produce nearly constant outputs. GAN-based approaches (BRITS w/ GAIL and NAOMI) hallucinate incorrect structures in incorrect locations. Our BDC Transformer also struggles in the middle of long gaps, but has the strongest performance overall, most closely reconstructing the ground-truth.}
\label{fig:allviz}
\end{center}
\vspace{-.6cm}
\end{figure*}


\hiddensection{Related Work} \label{sec:relatedwork}


Prior works are summarized in Table \ref{tbl:necessarycomp} and include: 1) Standard time-series imputation datasets and associated deep-learning methods; 2) Imputation approaches in mHealth Systems; and 3) Pulsative signal imputation research. PulseImpute provides the first comprehensive imputation benchmark for mHealth pulsative signals. 

\textbf{Standard Imputation Datasets}:
Prior time-series imputation work uses Traffic \cite{PEMSSF}, Air Quality \cite{kdd}, Billiard Ball Trajectory \cite{fragkiadaki2015billiards}, Sales \cite{deepmvi}, and other miscellaneous time-series modalities. Health-related imputation works ~\cite{hetvae,tar_brits,hr_metalearning,heartimp,sleep_mhealth} have benchmarked clinical record imputation~\cite{physionet2012, mimic} with non-pulsative signals (e.g. hourly body temperature). All of these prior datasets lack \emph{the high frequency and variable morphologies of pulsative waveforms} \cite{cstress, iranfar2021relearn}. For context, our 5-minute, 100 Hz waveforms are $\sim$600 times longer than the time-series data in the PhysioNet2012 dataset \cite{physionet2012}. 

Furthermore, many of these works \cite{brits, e2gan} simulate missingness by dropping independent time points at random, which is \emph{not representative of real-world mHealth missingness patterns}, as seen in Figure~\ref{fig:allviz}. State-of-the-art deep-learning time-series imputation methods \cite{brits, deepmvi, NAOMI} that were developed in these settings perform poorly in our PulseImpute Challenge (see Section~\ref{sec:results}). 

\textbf{Imputation in mHealth Systems}:
\cite{cstress_knn} develops a stress biomarker system with mHealth field data that handles real-world missingness for single-channel ECG signals. However, they do not quantify the effect of missingness and use a simple KNN multivariate imputation method. Additionally, their dataset is not publicly-available and lacks ground truth imputation values, so it cannot be used as a benchmark. \cite{iranfar2021relearn} also addresses stress detection in a real-world private dataset and only benchmarks simple imputation methods: multivariate iterative imputer, mean filling, last observation carried forward, or simply discarding instances with missingness. \cite{activity_knn} utilizes KNN multivariate imputation in an mHealth accelerometry dataset which is publicly-available but these signals are not pulsative. These representative works illustrate the challenges in addressing and benchmarking imputation methods in the context of real-world mHealth systems deployment.  




\textbf{Pulsative Signals Imputation}:
Prior work has focused on using non-deep-learning imputation methods \cite{varim,rpca,manimekalai2018knbp} on pulsative signals to address \emph{multi-channel} ECG imputation on publicly-available datasets. 
The key difference is that these methods can borrow information for imputation across ECG channels, a straight-forward task since the channels are highly-correlated (each channel measures the heart's electrical activity with respect to a different measurement axis). Multi-channel ECG recordings are routinely captured in clinical settings, but, in mHealth applications, where ECG is continuously measured with a wearable sensor, single-channel recording is the only practical approach. Therefore, we focus on \emph{single-channel} imputation. This requires the much more interesting and challenging task of borrowing information across time for imputation rather than across highly-correlated channels. Please see Appendix A1.2 for further discussion. 

\hiddensection{PulseImpute Challenge Description}
\label{sec:challenge}


We focus PulseImpute on the imputation of ECG and PPG signals, illustrated in Figure~\ref{fig:allviz}, because these widely-available pulsative signals are used in a wide range of mHealth and clinical tasks, such as monitoring atrial fibrillation \cite{hickey2017afib}, vascular aging \cite{castaneda2018ppg}, respiration rate \cite{hartmann2019ppgresp}, and stress \cite{cstress}.\footnote{We note that pulsative signals can arise in both mHealth and clinical settings, for example continuous waveform data may be captured in the ICU. A more detailed comparison of these settings is in Appendix A1.}
To quantify imputation performance, we simulate missing data by ablating samples, so that the imputed samples can be compared to the original ground-truth values. Prior imputation datasets have used relatively simple approaches to simulate missingness, typically by independently removing samples at random \cite{brits, e2gan, NAOMI}. We adopt two approaches for generating the block missingness patterns that characterize the mHealth domain. The first is extracting patterns of missingness from real-world mHealth studies \cite{chatterjee2020smokingopp, reiss2019ppgdalia}, illustrated in Figure~\ref{fig:allviz} and Appendix A3. The second is randomly selecting windows of samples of a fixed duration for ablation. Across experiments, we can vary the window duration to quantify the impact of the amount of missingness on algorithm performance. We impute each signal modality independently (i.e. univariate time series imputation) because this is important in practice for mHealth systems and leads to a more interesting and challenging task.

We generate training and testing sets in the PulseImpute Challenge by applying our missingness models to waveform data obtained from two existing clinical datasets: MIMIC-III Waveforms \cite{moody2020mimicWAVEFORM} (containing ECG and PPG signals) and PTB-XL \cite{ptbxl} (containing ECG signals). These datasets are large-scale, freely-available, and support a variety of downstream tasks for quantifying the impact of imputation performance on derived health markers. Specifically, MIMIC-III Waveforms supports \emph{heartbeat detection} using both ECG and PPG signals, in which the goal is to segment and localize individual heart beats. This is a core capability that supports a variety of widely-used mHealth markers such as heart rate variability \cite{hrv_info}. In the case of PTB-XL, the ECG waveforms have associated labels for \emph{rhythm, form, and diagnosis} classification tasks, such as atrial fibrillation detection, which are performed at the waveform level. These comprise a complex set of downstream clinical tasks which are impacted by imputation performance.

In summary, PulseImpute enables the evaluation of imputation algorithm performance at the signal level (each sample's reconstruction accuracy) and the downstream task level (quantifying the degradation in task accuracy due to imputation performance) for two widely-used pulsative signal types, ECG and PPG. In the next subsections, we describe the curation of our challenge datasets, our missingness models, and the performance metrics for our three downstream tasks: heartbeat detection with ECG and PPG, along with cardiac classification with ECG. In Section~\ref{sec:benchmarks}, we describe our suite of benchmark imputation methods. In Section \ref{sec:results}, we present empirical results that quantify the performance of SOTA methods on our novel challenge task and highlight directions for future research enabled by PulseImpute.

\hiddensubsection{ECG Imputation and Heartbeat Detection} \label{sec:ecghrmtask}
The goal of this task is to apply extracted real-world missingness patterns to ECG waveforms and formulate a downstream task of heartbeat detection. Imputation performance is assessed with reconstruction accuracy (signal level) and accuracy in detecting and localizing the ECG peaks corresponding to the heartbeats (task level).

\textbf{Dataset}: We have curated the largest clean public ECG waveform dataset available, containing 440,953 100 Hz 5-minute ECG waveforms from 32,930 patients. Our starting point was the raw ECG signals from the 6.7 TB MIMIC-III Waveforms dataset.\footnote{This is different from the MIMIC-III Clinical dataset, which is more commonly used, but which contains low frequency vitals data, not raw waveforms.} This dataset contains a variety of waveform data (e.g. ECG, PPG, etc.) and up to three ECG leads per patient. 
MIMIC’s unstructured nature with variable lead availability and imprecise electrode placements per recording \cite{gow_2021_mimicleadambigious} lends itself to a "union of leads" approach to dataset curation, in which we take each of the 
lead channels and add them separately to the dataset as individual univariate time series. This modifies the multivariate time-series into multiple univariate time-series. As previously discussed, univariate, single-channel recording is the norm for mHealth applications, and the inclusion of different leads in the union of leads dataset forces imputation methods to learn to borrow information across time to capture morphology due to each lead's distinctive shape. Additionally, while the most popular mHealth lead is lead I \cite{apple_2021, hall2020alivecor}, a wide range of lead configurations have been experimented with \cite{lai2020nonstandardecg, sprenger2022applewatchchestlead}, so this approach is useful for developing an imputation model that can generalize to different leads.

The key curation step was to filter out waveforms that were too noisy to support beat detection, while preserving those with irregular heart beat patterns. We used Welch's method \cite{WelchPeriodogram} and identified peaks in the periodogram which reveal the harmonics of the QRS complex. Tests on the peak distribution and spacing were used to identify clean ECG signals corresponding to typical as well as abnormal heart rhythms, while rejecting noisy samples. Further details can be found in Appendix A2.1. In contrast to our approach, prior works with MIMIC-III Waveform have used a random subset (1,000 2-minute ECG signals from 50 patients \cite{mimic_randomsubset}) of the data or a smaller matched subset that has corresponding clinical data (30,124 5-second ECG signals from 15,062 patients \cite{mimic_match}). We believe we are the first to preprocess the ECG MIMIC-III Waveform dataset in its entirety.

The resulting curated signals are very long, measuring 30,000 time points, which adds a level of complexity towards this challenge to have an emphasis on models that are scalable. For example, a transformer's self-attention mechanism is $O(n^2)$, and therefore, can't be applied naively. From an application's perspective, such long recordings are used because heart rate variability is most commonly measured in 5-minute intervals \cite{hrv_info}.

\textbf{Missingness}: We obtained \emph{extracted ECG mHealth missingness} patterns from our mHealth field study with 169 participants \cite{chatterjee2020smokingopp}. 
The missingness patterns are variable: most (69\%) of the missingness gaps are 3-9 seconds long, but some (2\%) of the gaps last more than a minute. Appendix A3.1 contains visualizations and further descriptions of the extracted missingness patterns. We have extracted 102,201 5-minute missingness masks, which capture the complex, real-life missingness patterns produced by wearable sensors in field conditions. 

\textbf{Downstream Task}: We use the Stationary Wavelet Transform peak detector \cite{peakdetector} to identify the sequence of peaks corresponding to individual heartbeats. Ground truth peaks are found by running the detector on the non-ablated signals. Peaks in the imputed signal are matched to the true peaks with a 50ms tolerance window~\cite{qin2017peakdetecttol50} to identify true vs false positives and define a detection problem (see Appendix A2.4 for details). We use the standard measures F1 score, precision, and recall, in order to quantify peak detection performance~\cite{liu2018f1, cai2020f1, detecttol}. 
95\% confidence intervals are generated by bootstrapping test samples with 1,000 iterations.


\hiddensubsection{PPG Imputation and Heartbeat Detection} \label{sec:ppghrmtask}
Analogous to Section~\ref{sec:ecghrmtask}, the goal of this task is to apply extracted real-world missingness patterns to PPG waveform data and formulate a downstream task of heartbeat detection in PPG.

\textbf{Dataset}:  We have curated the largest clean public PPG waveform dataset available, containing 151,738 100 Hz 5-minute waveforms from 18,210 patients.  We started with the raw PPG signals from MIMIC-III Waveforms, comprising of a variable-length univariate time-series for each participant cropped to be 5 minutes. To identify clean PPG waveforms and reject noisy signals, we used the approach from~\cite{vest2018toolbox} to perform beat segmentation and an ensemble averaging approach to identify a beat template for each waveform, which then is used to obtain a per-beat quality measure. Clean recordings had 95\% of the beats with quality higher than 0.5 (see Appendix A2.2 for details).


\textbf{Missingness}: As with ECG, we used a real-world mHealth PPG dataset, PPG-DaLiA~\cite{reiss2019ppgdalia}, to obtain \emph{extracted PPG mHealth missingness} patterns. We identified low quality beats using the method from~\cite{vest2018toolbox} and marked them as missing (see Appendix A2.3 for details). This resulted in 425 missingness masks which were used to ablate the curated clean PPG waveforms.


\textbf{Downstream Task}: Analogous to ECG, we used peak detection in PPG to identify individual beats (see Appendix A2.4 for details). Peaks in the clean waveforms provided ground truth for evaluating the imputed signals with F1 score, Precision, and Recall used to measure performance with 95\% confidence intervals generated from 1,000 bootstrapped iterations.

\hiddensubsection{ECG Imputation and Cardiac Pathophysiology Multi-label Classification} \label{sec:cardiactask}
This task focuses on quantifying the downstream impact of imputation on the challenging task of classifying cardiac disease conditions from ECG signals, by systematically assessing how varying percentages of missingness impact downstream performance.

\textbf{Dataset}: 
We utilize PTB-XL~\cite{ptbxl}, which is composed of 21,837 100 hz 10-second ECG waveforms from 18,885 patients, annotated with 71 labels that cover diagnostic, form, and rhythm categories of cardiac conditions. For each of the categories, the labels within them are not disjoint, resulting in a multi-label classification problem. We need to adapt the 12-lead ECG PTB-XL data and the SOTA multivariate xResNet1d classifier \cite{ptbxl_bench} to the univariate setting in order to create a downstream task. Since our goal is to assess imputation performance on classification,  the domain shifts associated with using different leads in a "union of leads" approach would be a confounding factor for the classifier. Our experiment design therefore uses a single lead, Lead I, the most common mHealth lead (e.g. Apple Watch and AliveCor \cite{apple_2021, hall2020alivecor}) and modifies the classifier to have an input channel size of 1. 



\textbf{Missingness}: We simulate two types of missingness patterns corresponding to long and short intervals of data loss. \emph{Extended Loss} ablates a random single contiguous set of samples as a percentage of the waveform duration. This models the most common source of missingness, sensor attachment issues, which comprise $\sim$85\% of total missingness \cite{arewethereyet}. In contrast, \emph{Transient Loss} models the sporadic loss of packets of samples due to communication failures or throttling of the data collection app~\cite{arewethereyet}. It is achieved by dividing the waveform into disjoint 50 ms blocks and sampling independently according to a fixed percentage of missingness to select blocks for ablation. The 50ms block size was selected to match standard packet sizes for mHealth data~\cite{cstress, packet62}. Extended and transient loss are both parameterized by a missingness percentage that controls (on average) the proportion of ablated samples in a waveform. In contrast to Section~\ref{sec:ecghrmtask}, our goal here is to characterize the impact of varying the percentage of missing data points at testing time on the reconstruction accuracy and downstream task performance. Imputation models are trained at a fixed 30\% missingness percentage (30\% was most common amount of missingness found in a 10 sec signal in our mHealth field study \cite{chatterjee2020smokingopp}). During testing, samples are ablated using percentages from 10\% to 50\% at a step size of 10\%, making it possible to quantify the effectiveness of imputation methods in generalizing to varying amounts of missingness at testing time.

\textbf{Downstream Task}: With \cite{ptbxl_bench}, three xResNet1d  \cite{xresnet} multi-label classification models for predicting diagnosis (e.g. WPW Syndrome), form (e.g. inverted T-waves), or rhythm (e.g. aFib) labels were trained on non-ablated data. Then, for all of the extended and transient missingness scenarios, after imputing a separate, held-out dataset with a given imputation method, each of the trained xResNet1d were evaluated on the imputed waveform to quantify the impact of imputation on clinical tasks that leverage the rhythm and morphology of the ECG waveforms. Classification results are measured using Macro-AUC, which is a common measure for multi-label classification under label imbalance \cite{spyromitros2011auc1, daniels2017auc2, jamthikar2022auc3} and has been theoretically proven to be optimized when the instance-wise margin is maximized \cite{wu2017aucproof}. The confidence intervals are generated identically as previously described.

\hiddensection{Benchmarks and Our Bottleneck Dilated Convolutional Self-Attention}
\label{sec:benchmarks}

\begin{figure*}[]
\begin{center}
\includegraphics[width=.95\textwidth]{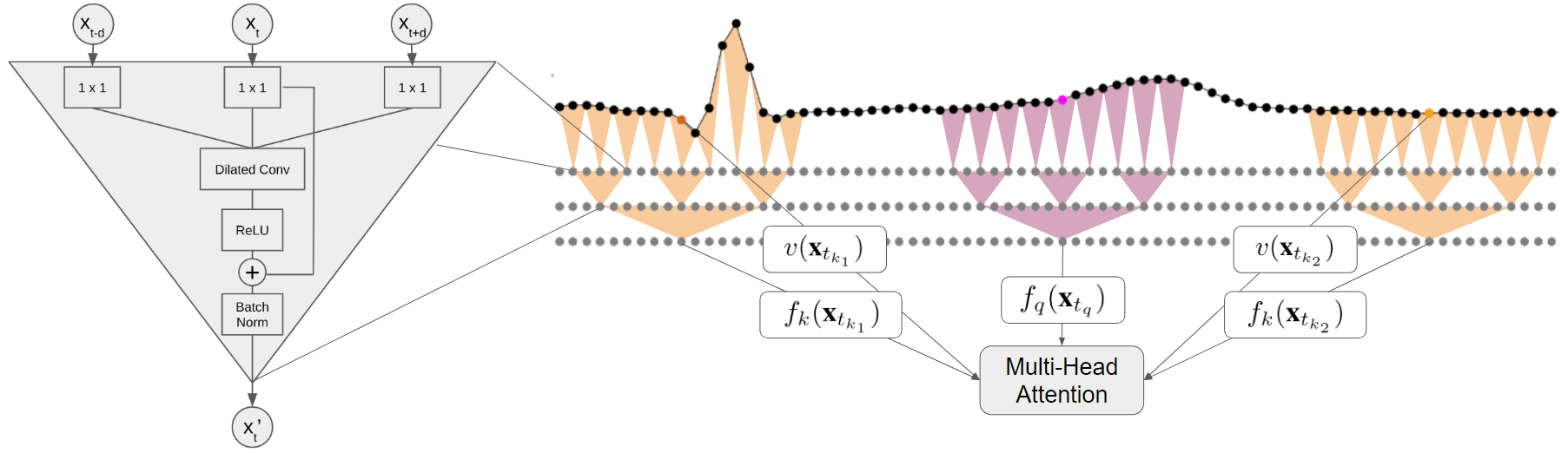}
\vspace{-.2cm}
\caption{Novel Bottleneck Dilated Convolutional (BDC) Self-Attention Architecture. The sequence of black dots on the right illustrates the ECG time series input to the module.
Each triangle on the right denotes a BDC block, shown in an expanded view on the left. The magenta dot denotes a query point for self-attention, while the orange dots denote representative queries. The query/key functions are composed of stacked blocks, denoted by the hierarchical structure of the colored triangles which illustrate the exponentially increasing dilation factor and receptive field used in the query/key functions. This enables efficient comparison of local context comprising 100s of samples. 
}
\label{fig:qkarch}
\end{center}
\vspace{-.6cm}
\end{figure*}


This is the first comprehensive study of mHealth pulsative signal imputation, and therefore there is a lack of prior baselines. We cover a range of classical methods to demonstrate baseline performance and use SOTA deep-learning methods from the general time-series imputation literature. In total, our benchmark suite includes the ten methods listed in Table~\ref{tbl:heartrateanal} with performance shown in Section \ref{sec:results}. 

Classical methods include mean filling, linear interpolation, and a Fast Fourier Transformer (FFT) imputer. Mean filling and linear interpolation are commonly used in mHealth \cite{le2018mean, dong2019lininterp} and help baseline the performance of more complex methods. We include an FFT imputer as a simple method that is able to utilize frequency information to exploit the quasiperiodic nature of the data \cite{rahman2015fft}. 

For deep-learning methods, we include DeepMVI \cite{deepmvi}, NAOMI~\cite{NAOMI}, BRITS w/ GAIL \cite{NAOMI}, BRITS \cite{brits}, Vanilla Transformer \cite{originaltransformer}, and Conv9 Transformer \cite{convattn}. DeepMVI is a transformer-based architecture that achieves SOTA in ten diverse real-world time-series datasets, ranging from air quality to sales \cite{deepmvi}. NAOMI develops a non-autoregressive approach paired with Generative Adversarial Imitation Learning (GAIL) \cite{gail}, and achieves SOTA performance for imputation tasks framed around trajectory modeling \cite{NAOMI}. NAOMI's backbone architecture is BRITS, a widely-used pure RNN imputation benchmark with a time-delayed loss propagation, that achieved SOTA in its benchmarked datasets \cite{brits}. BERT \cite{bert} used the vanilla transformer \cite{originaltransformer} with a masked language modeling imputation task for learning language representations, and the Conv9 Transformer \cite{convattn} was proposed to address the lack of local context while modeling time-series, which we will further discuss below.



Most of these methods were not designed to exploit the quasiperiodicity of our pulsative signals, so we anticipate that each of these methods would perform poorly in our setting. Therefore, we develop a transformer-based architecture that can provide a SOTA baseline for this domain. We claim that the pair-wise comparisons in the transformer's self-attention module are an attractive method for modeling the quasiperiodic dependency structure of pulsative signals.
In order to fully realize the potential of transformers in this setting, we identify three challenges that must be addressed: 1) local context, 2) permutation equivariance, and 3) quadratic complexity. This then motivates the development of the Bottleneck Dilated Convolutional (BDC) transformer baseline illustrated in Figure~\ref{fig:qkarch}. We now describe how our BDC Transformer addresses the three challenges.

\begin{figure}[]
\centering
\includegraphics[width=.65\textwidth]{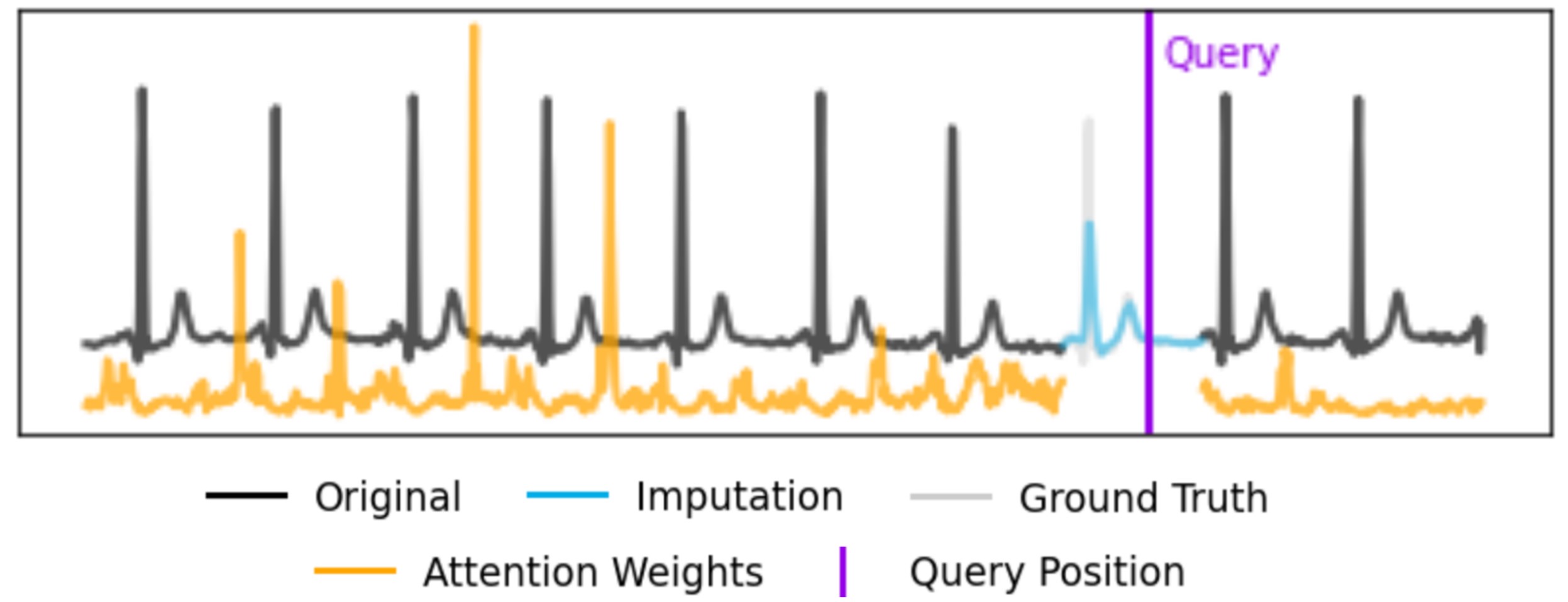}
\vspace{-.2cm}
\caption{Visualization of attention weights of BDC self-attention for a given query position. Instead of encoding a bias for time points close to the query, as done in prior work \cite{luo2015_grui,brits}, BDC attends to locations which are far from the query but similar in morphology, exploiting the quasiperiodicity. 
}
\label{fig:attnweights}
\vspace{-.2cm}
\end{figure}

\noindent\textbf{Local Context:} Transformers utilize self-attention, which we define below for a given query $\textbf{x}_q$:
\begin{align}
    A(\textbf{x}_q ) 
    & = \underset{\textbf{x}_{t_k}\in  S_{\textbf{x}_{t_k}}}{\textrm{Softmax}} \left( \exp(\frac{ \langle f_q( \textbf{x}_q ), f_k( \textbf{x}_k) \rangle}{ \sqrt{D}}) \right)
    \textbf{x}_{t_k} \textbf{W}_v, \label{eq:kernelsmoother}
\end{align}
where $f_q( \textbf{x}_{t_q} ) =  \textbf{x}_{t_q} \textbf{W}_q $, $f_k( \textbf{x}_{t_k} ) =  \textbf{x}_{t_k} \textbf{W}_k $,  and $\textbf{W}_{\{q/k/v\}} \in \mathcal{R}^{D_x \times D}$. $ S_{\textbf{x}_{t_k}}$ is the set of all key time points, and $\textbf{x}_{t}$ is the $t$th row of $\textbf{X} \in \mathbb{R}^{T \times D_x}$, where $D_x$ is the dimensionality of the time-series.
In NLP applications, each input word token has intrinsic semantic information which allows for meaningful direct comparison via self-attention in Eq.~\ref{eq:kernelsmoother}. In contrast, in order to meaningfully compare two timepoints in a time-series, it is necessary to utilize the local signal context around the queries and keys. This can be accomplished via convolutional self-attention \cite{convattn}, which models the query and key function as a convolution, which we demonstrate for $f_q$ below:
\begin{align}
    f_{q}(\textbf{x}_{t_q})&= (\textbf{X} \star h) [t_q] = 
    \sum_{s=-\infty}^\infty \textbf{x}_{s} h_{t_q + s}  
    \text{ where } h_u = 
    \begin{cases} 
      W_u^{(q)}  & \lfloor \frac{1-i}{2} \rfloor \leq u \leq \lfloor \frac{i-1}{2}  \rfloor \\
      0 &  \text{elsewhere}
   \end{cases},  \label{eq:convattn} 
\end{align} 
where $i> 1$ is the filter size and with $W_u^{(q)} \in \mathbb{R}^{D_x \times D}$ as the $u$th row of $W^{(q)} \in \mathbb{R}^{i \times D_x \times D}$. The original conv transformer implementation \cite{convattn} had a single convolution with a small filter size of 9. The key for this approach to be effective in our pulsative signal setting is to achieve a sufficiently large receptive field (RF) to achieve subsequence comparisons between patterns lasting for 100s of time points, while maintaining computational efficiency. We use stacked dilated convolutions\comment{~\cite{yu2015dilated}} with bottleneck 1x1 layers in our novel BDC architecture illustrated in Figure \ref{fig:qkarch}. The bottleneck reduces dimensionality, allowing us to stack filters with exponentially increasing dilation factors, thereby exponentially and efficiently increasing the RF. We can see empirically in Table \ref{tbl:paramfixed}, that while both BDC and conv transformers expand receptive field (RF), only BDC improves performance relative to the vanilla self-attention after controlling for parameter count.


Prior time-series imputation architectures such as BRITS \cite{brits}, encode a bias for time points that are close to the query. This is not effective for pulsative signals, where temporal locations which are far from the query can be similar in morphology, and thus useful for imputation. Our BDC Transformer is able to exploit this, as shown by the learned attention weights illustrated in Figure \ref{fig:attnweights}. 

\noindent\textbf{Permutation Equivariance:} A permutation of a transformer's inputs results in a corresponding permutation of it's outputs without any change in values~\cite{permequiv}. This is addressed in the original transformer's formulation via an additive positional encoding \cite{originaltransformer}, but this 
does not have a good inductive bias in our setting. The absolute position relative to the start of a pulsative signal is not meaningful, due to the arbitrary start-time at which sensors begin recording \cite{hrv_info}. Even if the signals were initially aligned, due to within-subject and between-subject phase variance stemming from the heart rate variance phenomena \cite{hrv_info}, the relative position of specific waveform shapes will vary. 

\begin{table}[]
\centering
\small
\begin{tabular}{|c|c|c|c|}
\hline
Models w/ Params fixed at ${\sim}2.6$ mil & Van Trans RF=1 & Conv Trans RF=9 & \textbf{BDC Trans RF=883} \\ \hline
MSE  $\downarrow$  & 0.0177 & 0.0231 & \textbf{0.0123} \\ \hline
\end{tabular}%
\vspace{.1cm}
\caption{Comparison of models with parameters fixed at ${\sim}2.6$ mil. With its stacked dilated convolutions, BDC is able to efficiently increase RF and improve performance relative to vanilla self-attention. }
\label{tbl:paramfixed}
\end{table}

\begin{table}[]
\centering
\small
\begin{tabular}{|c|c|c|c|c|c|c|c|}
\hline
Model &  \textbf{BDC Trans} & PE+BDC Trans & Conv Trans & PE+Conv Trans \\ \hline
MSE  $\downarrow$    & \textbf{0.0118}     & 0.0121          & 0.0223    & 0.0225         \\ \hline
\end{tabular}%
\vspace{.1cm}
\caption{Positional encoding (PE) slightly degrades performance for both BDC and Conv Transformer. } \label{tbl:RFandPE}
\vspace{-.6cm}
\end{table}

Now, we note that from the conv self-attention formulation in Eq. \ref{eq:convattn}, one can see that conv and our extension, BDC, self-attention, are no longer permutation equivariant because the calculation at each position depends on its neighbors. An additional additive positional encoding would perturb the original signal, potentially rendering the imputation task more difficult. Indeed, we empirically demonstrate in Table \ref{tbl:RFandPE} that including an additive positional encoding \cite{originaltransformer} degrades model performance. Therefore, we design our approach around the BDC self-attention without a positional encoding, because of its strong inductive bias in its ability to encode local context, while also breaking permutation equivariance.


\noindent\textbf{Quadratic Complexity:} Transformers have quadratic time and space complexity for self-attention that limits applications to long sequences \cite{tai2020effecienttrans}. The Longformer~\cite{longformer} dilated sliding window attention
restricts key range in self-attention, $ S_{\textbf{x}_{t_k}}$, without modifying query/key functions, allowing it to be easily combined with our BDC self-attention. We use this longformer variant for the transformer models in the 5-minute-long (30,000 time points) time-series used in the heartbeat detection tasks. 

\hiddensection{Results}\label{sec:results}

We now present comprehensive results for our baseline models on all of the PulseImpute Challenge tasks, organized by downstream task as described in Section~\ref{sec:challenge}. See Appendix A4 and our code repository on implementation details for reproducibility.




\begin{table}[]
\centering
\resizebox{\columnwidth}{!}{%
\begin{tabular}{@{}lcccccccc@{}}
\toprule
\multicolumn{1}{l}{}  & \multicolumn{4}{c}{ECG Imputation and Heartbeat Detection} & \multicolumn{4}{c}{PPG Imputation and Heartbeat Detection} \\ \cmidrule(l){2-5}  \cmidrule(l){6-9} 
\textbf{Models} & $\downarrow$ \textbf{MSE} & $\uparrow$ \textbf{F1} & $\uparrow$ \textbf{Prec} & $\uparrow$ \textbf{Sens} & $\downarrow$ \textbf{MSE} & $\uparrow$  \textbf{F1} & $\uparrow$  \textbf{Prec} & $\uparrow$  \textbf{Sens} \\ \midrule 
Mean Filling & .0278 $\pm$ .00019 & .01 $\pm$ .000 & .60 $\pm$ .000 & .00 $\pm$ .007 & .0971 $\pm$ .00123 & NaN & NaN & 0 $\pm$ 0 \\

Lin Interp & .0467 $\pm$ .00046 & .01 $\pm$ .000 & .62 $\pm$ .000 & .00 $\pm$ .009 & .1393 $\pm$ .00073 & NaN & NaN & 0 $\pm$ 0 \\

FFT \cite{rahman2015fft} & .0350 $\pm$ .00024 & .09 $\pm$ .001 & .07 $\pm$ .001 & .16 $\pm$ .001 & .1449 $\pm$ .00120 & .10 $\pm$ .001 & .07 $\pm$ .001 & .16 $\pm$ .001 \\

BRITS \cite{brits} & .0445 $\pm$ .00068 & .01 $\pm$ .000 & .32 $\pm$ .000 & .01 $\pm$ .001 & .1064 $\pm$ .00068 & .01 $\pm$ .000 & .03 $\pm$ .000 & .01 $\pm$ .001 \\

BRITS w/ GAIL \cite{gail} & .0571 $\pm$ .00068 & .05 $\pm$ .001 & .08 $\pm$ .001 & .03 $\pm$ .003 & .1102 $\pm$ .00068 & .07 $\pm$ .001 & .29 $\pm$ .001 & .04 $\pm$ .003 \\

NAOMI \cite{NAOMI}  & .0392  $\pm$ .00061 & .05 $\pm$ .001 & .13 $\pm$ .001 & .03 $\pm$ .001 & .0856 $\pm$ .00061 & .09 $\pm$ .001 & .09 $\pm$ .001 & .10 $\pm$ .001 \\

DeepMVI \cite{deepmvi}  & .0276 $\pm$ .00019 & .05 $\pm$ .000 & .49 $\pm$ .000 & .02 $\pm$ .005 & .0802 $\pm$ .00061 & .25 $\pm$ .001 & .31 $\pm$ .001 & .21 $\pm$ .002 \\

Vanilla Trans  \cite{originaltransformer}   & .0368 $\pm$ .00021 & .02 $\pm$ .001 & .29 $\pm$ .000 & .01 $\pm$ .005 & .0967 $\pm$ .00065 & .13 $\pm$ .001 & .15 $\pm$ .001 & .12 $\pm$ .001 \\

Conv9 Trans \cite{convattn}  & .0299 $\pm$ .00021 & .01 $\pm$ .000 & .47 $\pm$ .000 & .00 $\pm$ .007 & .0805 $\pm$ .00056 & .20 $\pm$ .001 & .20 $\pm$ .001 & .19 $\pm$ .001 \\

BDC Trans (ours)  & \textbf{.0194} $\pm$ .00017 & \textbf{.64} $\pm$ .003 & \textbf{.83} $\pm$ .003 & \textbf{.52} $\pm$ .002 & \textbf{.0137} $\pm$ .00020 & \textbf{.81} $\pm$ .003 & \textbf{.79} $\pm$ .003 & \textbf{.83} $\pm$ .003\\ \bottomrule
\end{tabular}%
}
\vspace{.1cm}
\caption{ECG and PPG imputation and heartbeat detection results using extracted mHealth missingness patterns (see Sec.~\ref{sec:ecghrmtask} and \ref{sec:ppghrmtask}). Measures are for reconstruction performance (MSE) and Heartbeat Detection accuracy (F1 Score, Precision, and Sensitivity) with 95\% Confidence Intervals.  }


\label{tbl:heartrateanal}
\vspace{-.6cm}
\end{table}

\noindent\textbf{ECG Imputation and Heartbeat Detection:} All prior time-series imputation models perform poorly on the long ECG time-series (30,000 time points) with complex ECG missingness patterns, which can be seen in Table \ref{tbl:heartrateanal} and the ECG column in Figure \ref{fig:allviz}. BRITS w/ GAIL and NAOMI hallucinate realistic ECG patterns but do not match the ground-truth. BRITS
fails to effectively impute over longer gaps. FFT and our BDC model have the best imputation performance, and can reconstruct the rhythm of the missing ECG peaks, as shown in the peak detection statistics, with BDC easily having the best MSE and F1 score overall, at 0.0194 and 0.64, respectively. BDC transformer can effectively capture the extended local context in comparison to other transformer models (e.g. DeepMVI, Conv9, Vanilla), reconstructing realistic ECG signals, reminiscent of the ground-truth.
However, as seen in the further visualizations in Appendix A5.1, none of the models are able to effectively impute over longer missingness gaps that can be up to one minute.


\noindent\textbf{PPG Imputation and Heartbeat Detection:} PPG is morphologically simpler than ECG (see Figure \ref{fig:allviz}), and most methods perform better with respect to their F1 score. However, as seen in Figure \ref{fig:allviz}, methods such as DeepMVI and BRITS can only impute values near observed data. 
FFT and BDC demonstrate that exploiting quasiperiodicity is useful across signal modalities, and BDC achieves the best overall results as seen in Table \ref{tbl:heartrateanal}. Further visualizations can be found in Appendix A5.2.



\begin{figure}[!t]
\begin{center}
\includegraphics[width=.85\textwidth]{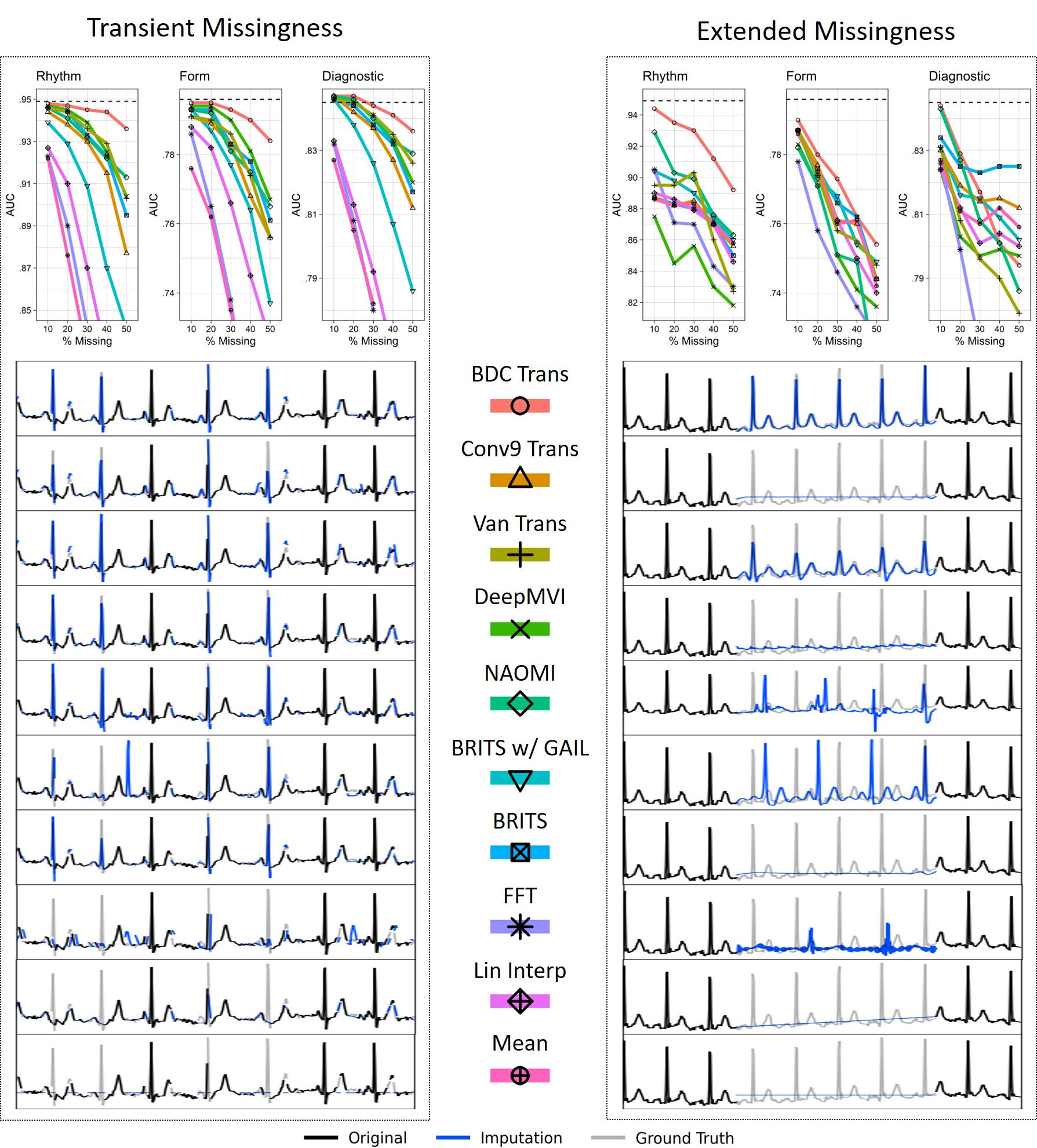}
\caption{Cardiac Classification in ECG Results for Transient and Extended Loss on Rhythm, Form, and Diagnosis label groups. For each label category, a cardiac classifier was trained and tested on complete data (top, illustrated by dashed line). The trained model was then evaluated on 
imputed test data (for five levels of missingness from 10\% to 50\%) produced by each baseline, yielding the Macro-AUC curves (top). Six seconds of representative imputation results for the 30\% missingness test case are plotted (below). The Extended Loss setting proved to be more challenging for all models.}
\label{fig:ptbxlclassviz}
\end{center}
\vspace{-.5cm}
\end{figure}

\noindent\textbf{ECG Imputation and Cardiac Classification:} The imputation visualizations and the downstream results in Figure \ref{fig:ptbxlclassviz} show that in the transient loss setting, many models outperform mean imputation in mimicking the Rhythm, Form, and Diagnosis features present in the original waveform, as reflected by their downstream Macro-AUC results for each category. Here missingness gaps are shorter and there is less need to learn long-term dependencies across those gaps. BRITS and Vanilla Transformer, imputation methods that were originally trained and designed for very short missingness gaps \cite{brits, bert}, perform well in this setting. FFT imputation performs poorly in this setting, but our BDC model has the best MSE and does the best in reconstructing rhythm, form, and diagnosis characteristics.

In the more challenging extended loss setting, imputation performance drops across all models, resulting in poor downstream performance. The GAN-based methods and FFT have poor performance, while the BDC model has the best MSE and the best performance in reconstructing rhythm and form features. 
However, in the diagnostic category, performance is poor. We hypothesize that this may be tied to BDC transformer reconstructing much shorter R peaks in the ECG signal under high missingness percentages, which can be seen in Figure A13 in the Appendix. Some diagnostic labels are dependent on R peak height (e.g. LAFB is diagnosed with tall R waves \cite{larkin_buttner_2021}), and thus will be adversely affected. We hypothesize that the minimalistic imputations produced by BRITS, Conv9, and mean filling (see Figure 3 and A13) fare better because there is less misleading signal information present, suggesting that the current imputation SOTA is inadequate for this challenging task. Full tabulated results with confidence intervals and extra visualizations can be found in Appendix A5.3.


\hiddensection{Discussion}

\noindent\textbf{Future Work and Limitations:} 
Our BDC architecture provides a useful starting point for exploiting quasiperiodicity. However, the visualizations in Figure~\ref{fig:allviz} (and in Appendix A5), demonstrate that all existing methods are unable to impute over missingness gaps lasting up to a minute in the heartbeat detection tasks. Additionally, all methods are far from the upper-limit of performance in each label group for Cardiac Classification in the extended loss setting. A key challenge is to improve imputation in the middle of long missingness gaps, which might benefit from a generative modeling approach.

A potential benefit of mHealth is the ability to analyze an individual's health-related behaviors so as to improve their health outcomes. Future work could include benchmarking for personalized models, similar to the approach proposed by \cite{deng2021deep}. Additionally, none of our benchmarked methods explicitly model the imputation uncertainty, and we plan to explore related architectures with uncertainty modeling such as HeTVAE \cite{hetvae}. Another potential area is explainability of such imputation models.
We note that the theoretical missingness model for this work is MCAR, as the missingness is independent of the waveforms that they are applied to \cite{mack2018mcar}. This was necessary to obtain ground-truth imputation targets, but future work should investigate the inclusion of MAR and MNAR missingness.

\textbf{Societal Impacts}:  We anticipate our work to have positive societal benefits by enabling researchers to address one of the most common issues found in mHealth, accelerating the field forward to enable individuals to live healthier lives. As with all ML challenges, there may be a negative environmental impact due to increased computational usage of researchers working on this challenge.

\hiddensection{Conclusion}
We introduced \emph{PulseImpute}, a novel imputation challenge for pulsative mHealth signals. We curated a set of ECG and PPG datasets with realistic mHealth missingness patterns and relevant downstream tasks. Our comprehensive set of baselines includes a novel Bottleneck Dilated Convolutional (BDC) transformer architecture that is able to exploit the quasiperiodicity present in our data and defines the SOTA. At the same time, our findings demonstrate that previous existing methods fail to achieve high performance, pointing out the need for additional research. PulseImpute addresses a significant gap in mHealth pulsative signal imputation, providing the first large-scale reproducible framework for the machine-learning community to engage in this unique challenge.


\newpage
\section*{Acknowledgements}
This work is supported in part by NIH P41-EB028242-01A1, NIH 7-R01-MD010362-03, and the National Science Foundation Graduate Research Fellowship under Grant No. DGE-2039655. Any opinion, findings, and conclusions or recommendations expressed in this material are those of the authors and do not necessarily reflect the views of the National Science Foundation.

\printbibliography


\newpage 
\appendix 
\renewcommand{\thesection}{\Alph{section}}
\begin{refsection}

\makeatletter

\renewcommand{\thesection}{A\arabic{section}}

\makeatletter

\renewcommand\thefigure{A\arabic{figure}}
\renewcommand\thetable{A\arabic{table}}

\title{Appendix for PulseImpute}
\author{}

\maketitle \vspace{-2cm}
\renewcommand{\contentsname}{Appendix Table of Contents}
\tableofcontents

\vspace{2cm}
\newpage 
\section{mHealth versus Clinical Pulsative Waveforms}
In our proposed challenge, we draw from clinical pulsative waveform datasets to mimic mHealth pulsative waveforms, and in this section, we provide additional justification for this approach.

\subsection{What is the rationale for constructing a dataset for mHealth signal imputation from clinical patients?} 
While there are differences between clinical pulsative signals collected in a hospital setting and mHealth pulsative signals collected in the field, this was a necessary approach due to the scarcity of large, publicly-available mHealth datasets (e.g. PPG-DaLiA, an mHealth dataset, has 15 subjects whereas our curated MIMIC-III PPG dataset, derived from a clinical dataset, has 18,210 subjects). Then, we can mimic real-world mHealth settings by applying realistic patterns of mHealth missingness while using the original ablated samples as ground truth, making it possible to quantify and visualize the accuracy of imputation. 

\subsection{What are the differences in how the sensors collect pulsative signals?} 
An ECG lead is the measure of the heart's activity along a specific axis, by measuring the electrical voltage difference across given electrodes. Because in a clinical hospital setting the patients are stationary, it is simple to attach multiple electrodes onto the patient for diagnosis or monitoring purposes, allowing for multiple ECG leads to be recorded. However, in an mHealth field setting, ECG is recorded using wearables, such as a smart watch \cite{apple_2021} or a band \cite{ertin2011autosense}, on a user who may be constantly moving. Therefore, to prevent creating an unacceptable burden, single-lead recording is typically the only acceptable approach. 
This difference in leads used is why during curation, we treat each lead as a separate waveform, and propose a univariate imputation problem rather than a multivariate one. In addition to this, because both mHealth sensors and clinical sensors will calculate ECG with the same collection procedure of measuring changes in voltage between electrodes, differences in clean signals will reduce down to the lead measured, rather than mHealth versus clinical setting. Previous work \cite{gropler2018mhealthvsclinical, kathleen2017mhealthvsclinical2} has demonstrated that mHealth ECG sensors are able to record clinically-accurate single-lead ECG tracings in both healthy subjects and subjects with underlying cardiac disease or rhythm abnormalities.

The PPG sensing modality may also vary in mHealth versus clinical setting as well. In clinical hospital settings, the device is clipped to the stationary patient's finger so the device is more stable and gives a higher data quality \cite{nardelli2020ppgfingervswrist}. However, in mHealth, PPG signals are typically collected on a watch, so if the device is not strapped well, there will be more noise and missingness resulting from movement \cite{nardelli2020ppgfingervswrist}. While there are some differences in shape between wrist mHealth PPG and finger clinical PPG \cite{rajala2018ppgfingervswristshape}, they both model the same paradigms, such as morphological-based ones (e.g. Pulse Arrival Time \cite{rajala2018ppgfingervswristshape}) and rhythm-based ones (e.g. Heart Rate Variability \cite{nardelli2020ppgfingervswrist}), and these works demonstrate how PPG signals collected in the mHealth setting may be adapted to be used for clinical marker calculations that originally designed for clinical PPG signals \cite{rajala2018ppgfingervswristshape,nardelli2020ppgfingervswrist}. This suggests that domain gap issues between clinical and mHealth settings, while they exist, may not be not a major obstacle.

\subsection{How do the patient populations differ in these two settings?}
Generally, patients in a clinical setting are in a worse health condition than users in a mHealth setting. mHealth technology has a large consumer base and is generally used by individuals for maintenance of healthy behaviors. In the hospital, patients may be in the ICU with ECG/PPG sensors to monitor their already-poor health condition. Therefore, clinical hospital signals will be more variable than those originating from healthy individuals due to the diverse set of diseases that may affect cardio-pulmonary systems and the pulsative signals that measure them. However, this does not appear to be a source of limitations, as we are presenting a more interesting challenge for the ML community to tackle. ML methods must rely on learning to impute missing signals based on the signal that is present, rather than learning to create a general-purpose template that may be applied for most individuals.

\subsection{How does clinical missingness differ from mHealth missingness?}
There are similarities in missingness patterns across the mHealth and clinical domains. For example, with respect to participant compliance, both clinical patients and mHealth users can remove sensors, resulting in contiguous blocks of missing data. Likewise, participant movement in both contexts can result in artifacts (e.g. tugging at a finger-mounted sensor in the hospital, vs adjusting the strap of a mHealth wearable which becomes uncomfortable). At the same time, the mHealth environment is more challenging for data capture and may experience more missingness overall.

However, we would like to clarify that comparisons between clinical and mHealth missingness do not affect our findings and experimental design because:
\begin{enumerate}
    \item We do not make any claims about the suitability of our approach for addressing the issue of clinical missingness.
    \item There is no clinical missingness present in our benchmark dataset
\end{enumerate}

Our contribution is on introducing a benchmarking suite for pulsative signals with realistic mHealth missingness, and our data curation process (described in Sections 3.1, 3.2, A2.1, A2.2) ensured that signals with clinical missingness were removed from the dataset (e.g. a waveform experiencing missingness due to the sensor being away from the body would fail the peak detection tests found in our preprocessing process). 

\newpage 

\section{Curation of MIMIC-III Waveform and Heartbeat Detection Task Details} \label{sec:curation} 
For each of these curations, we intentionally utilize an aggressive filtration method to ensure that we have clean signals, as the sheer size of MIMIC-III Waveforms allows us to filter out many unsuitable signals and still curate the largest ECG/PPG waveform dataset. There are multiple reasons why we would like to curate a clean dataset version of MIMIC-III Waveform. The first is that we would like our challenge to focus on the novel idea of exploiting quasi-periodic structure in this unique time-series domain to achieve strong performance. Therefore, we want signals in which the structure is clean. 
Secondly, we want to match the quality level of other datasets such as PTB-XL, in which 77.01\% of the signal data are of highest assessed quality \cite{ptbxl}. This matching enables researchers to combine datasets in the future, and allows for the possibility of transfer learning approaches, potentially leveraging self-supervised representation learning approaches. 

\subsection{MIMIC-III ECG Curation}
MIMIC-III Waveform has 4,799,017 ECG signal files, which we curate down to 440,953 clean ECG signal files. For a given ECG Signal in MIMIC-III Waveform \cite{moody2020mimicWAVEFORM} we...
\begin{enumerate}[nosep]
    \item Resample the waveform from 125 Hz to 100 Hz and conduct linear interpolation to fill in NA values
    \item Utilize Welch's method \cite{WelchPeriodogram} to obtain the periodogram and conduct peak detection on the periodogram with a strict minimum peak distance requirement. In a clean ECG signal, regularly spaced peaks in the periodogram correspond to the harmonics of a QRS complex, especially those in the upper frequency bands \cite{bashar2019ecgnoisespectrogram}. Therefore, if the detected peaks are regularly spaced and there is a peak detected at $>10$ Hz, then the ECG signal is marked as clean, and we skip to step 4. If the peaks are not too irregularly spaced, then we move to step 3 for another chance for the signal to pass the quality check. See below for examples of ECG signals with their associated periodogram. The top demonstrates a clean ECG signal with regularly spaced peaks in its periodogram and detected peaks past 10 Hz, and the bottom represents a noisy ECG signal with a periodogram with irregularly spaced peaks and no significant peaks detected past 10 Hz.
    \\
    \includegraphics[width=\textwidth]{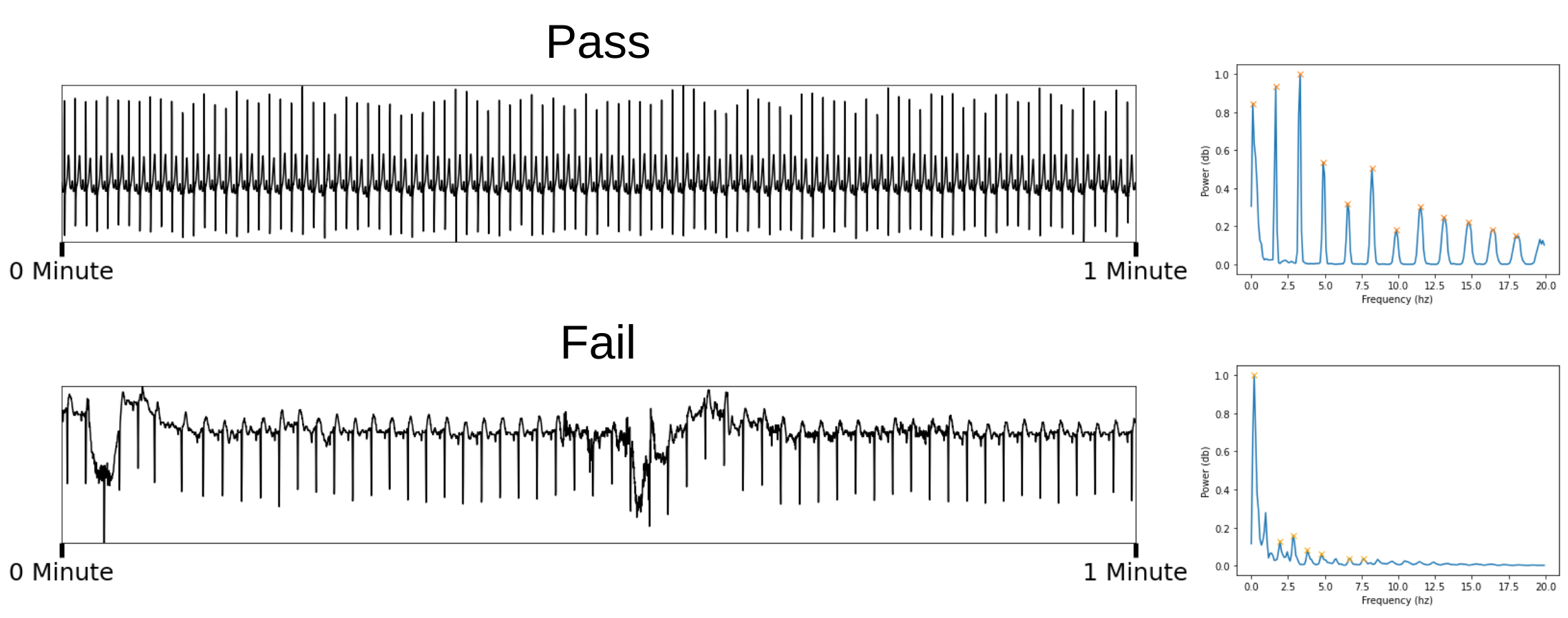}
    \item A new peak detection on the periodogram is conducted with a relaxed minimum peak distance requirement, and new peaks are compared to old peaks. These new peaks are designated by the red x in the below plots. If number of new peaks is not too high or if the new peaks are far away from the old peaks, then the signal is marked clean, and we move to step 4. This allows for non-normal heart rhythms in which the heart rate fluctuates to pass. These examples below demonstrate heart rate irregularities (top) and minor baseline drift noise (bottom).   \\
    \includegraphics[width=\textwidth]{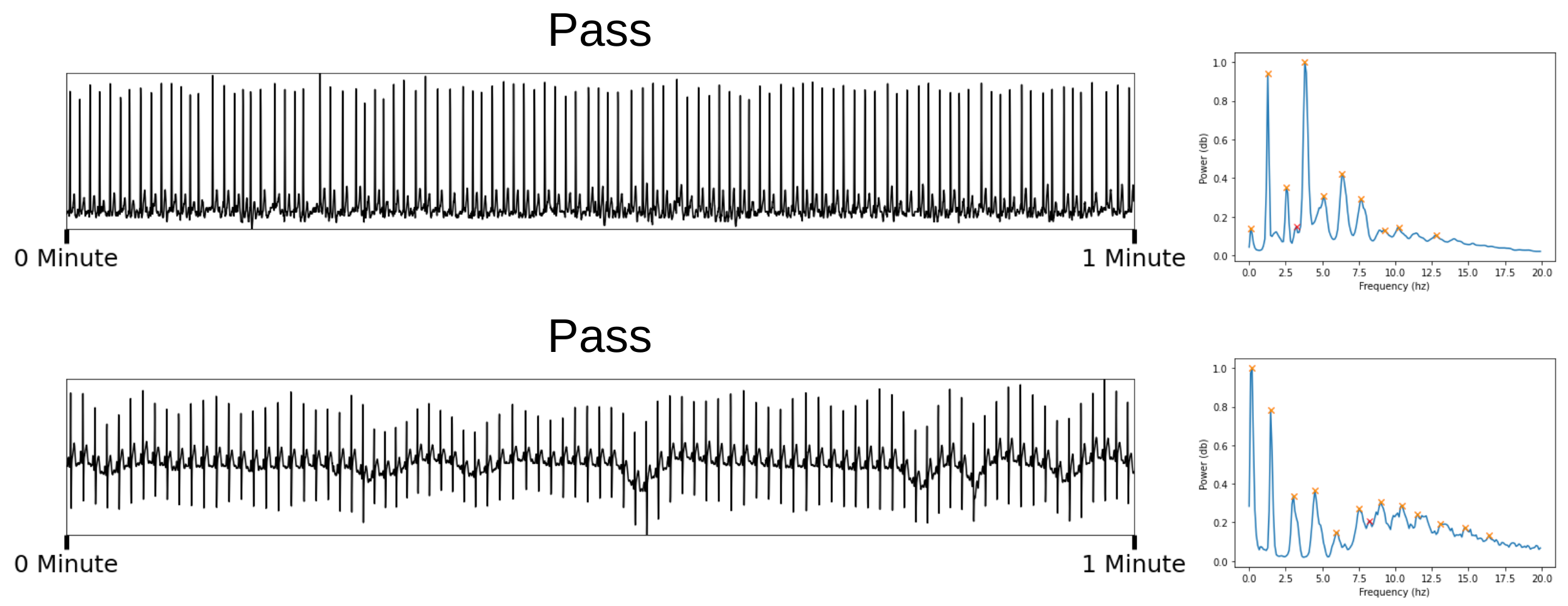}
    \item For all signals marked clean, we sample a 5-minute segment, run an ECG peak detection algorithm, and if HR is within acceptable physiological range, then the \emph{ECG signal is accepted}. 
\end{enumerate}

\subsection{MIMIC-III PPG Curation}
MIMIC-III Waveform has 3,162,804 PPG signal files, which we curate down to 151,738 clean PPG signal files. For a given PPG Signal in MIMIC-III Waveform \cite{moody2020mimicWAVEFORM} we...
\begin{enumerate}[nosep]
    \item Sample a 5-minute segment and resample the waveform from 125 Hz to 100 Hz
    \item Segment the waveform into each beat with a peak detection algorithm 
    and extract PPG beat template with ensemble averaging-based approach \cite{vest2018toolbox} . If template is failed to be found, then discard the signal.
    \item Calculate the DTW-based quality metric (bounded between 0 and 1) for each beat. This is done by using DTW to make both the template and the beat the same length and calculating the correlation coefficient. If the correlation is negative, the similarity is clamped to zero.
    \item If 95\% of the beats have a quality greater than 0.5, the \emph{PPG signal is accepted}. See below for examples of accepted and rejected PPG signals \\
    \includegraphics[width=.7\textwidth]{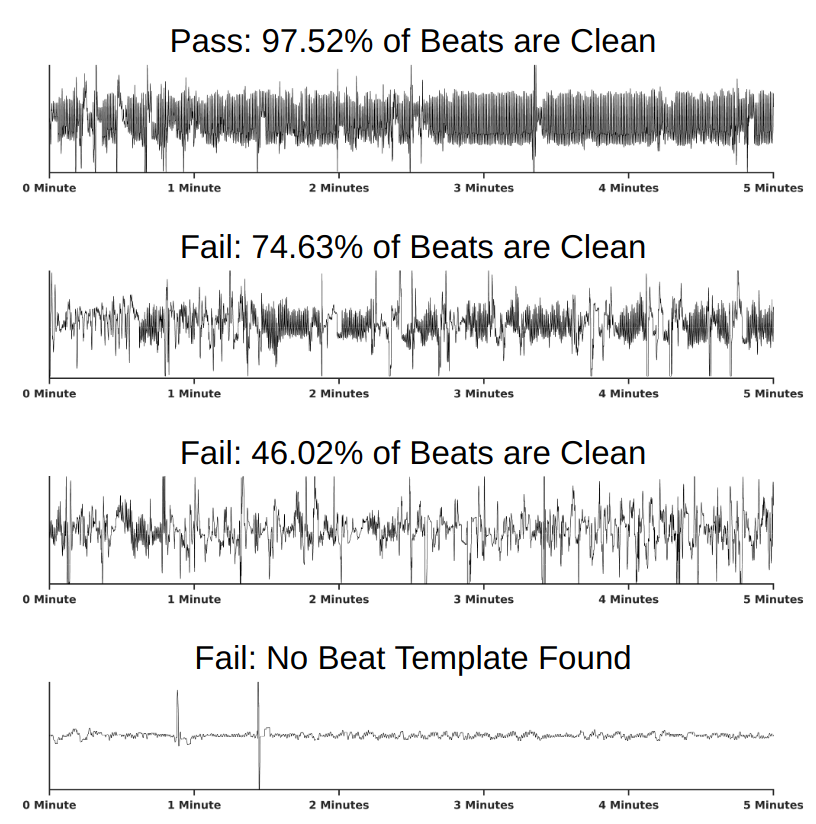}
\end{enumerate}

\subsection{PPG mHealth Missingness Extraction}
To generate ECG mHealth Missingness patterns, the Autosense \cite{ertin2011autosense} device in our mHealth field study \cite{chatterjee2020smokingopp} used an ECG data quality assessment algorithm \cite{md2k_2019} to detect noise and missingness. However, this Autosense device does not record PPG signals, and thus we do not have access to PPG mHealth missingness patterns. We cannot use ECG mHealth missingness patterns to model missingness in PPG because PPG signals have different missingness structure. One reason for this is that they are collected with different sensor attachments: ECG signals can be collected on a chest band, as is done in Autosense, whereas PPG signals are typically collected with a sensor on the wrist with a smartwatch. Therefore, we seek to extract missingness patterns from the public mHealth PPG dataset, PPG-DaLiA \cite{reiss2019ppgdalia}, with the procedure outlined below. 

For a given PPG Signal in PPG-DaLiA \cite{reiss2019ppgdalia} we...
\begin{enumerate}
    \item Resample the waveform from 64 Hz to 100 Hz
    \item Segment the waveform into each beat with PPG-DaLiA's provided ground-truth peaks and extract PPG beat template with ensemble averaging \cite{vest2018toolbox}.
    \item Calculate DTW-based quality metric (bounded between 0 and 1)
    \item Create a binary time-series by marking segments with quality $< .5$  as missing (0) and segments with quality metric $\geq .5$ as not-missing (1).
    \item Split the binary time-series into 5 minute segments to serve as PPG mHealth missingness patterns.
\end{enumerate}

\subsection{Heartbeat Detection via Peak Detection in ECG/PPG} 
Here we detail the peak detection procedure used for the downstream task of ECG/PPG Heartbeat detection on the curated MIMIC-III waveform datasets. Peak detection is essential for segmenting and localizing individual heart beats, which is a core capability that supports a variety of widely-used mHealth markers such as heart rate (HR) and heart rate variability (HRV). Below are the formulations of each of the metrics that we use in this task.

\begin{align*}
    \text{Sens} = \frac{TP}{TP + FN} \text{\ \ \ Prec} = \frac{TP}{TP + FP} \text{\ \ \ F1} = \frac{2 * \textrm{Prec} * \textrm{Sens}}{\textrm{Prec} + \textrm{Sens}}
\end{align*}

Given a peak that was identified from the imputation, we center a 50 ms window around this peak. If there was a peak originally in this window before being ablated for imputation, then this is a \emph{True Positive}. If there was no peak originally in this window, this is a \emph{False Positive}. \emph{False Negatives} are peaks that were in the original signal that were ablated, but were not detected in the reconstructed imputation with this procedure. The peak detection procedures were Stationary Wavelet Transform peak detector from~\cite{peakdetector}  for ECG signals and a neighbor comparison with threshold and peak prominence filters for PPG signals.


\newpage 
\section{mHealth Missingness Visualizations for ECG and PPG}

\subsection{ECG Extracted mHealth Missingness} 

\begin{figure}[!htbp]
\centering
\includegraphics[width=\textwidth]{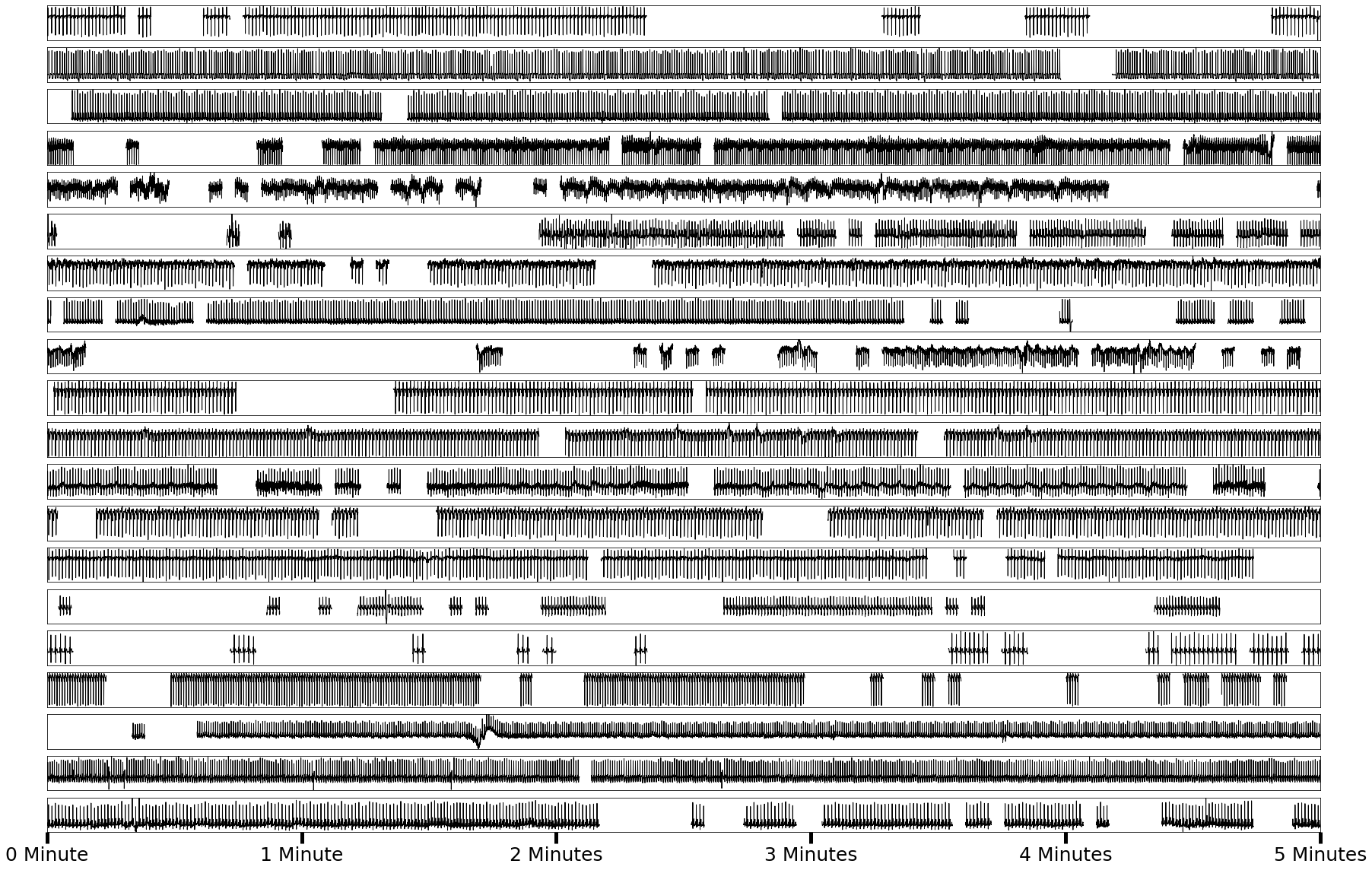}
\caption{ Various extracted ECG mHealth signal missingness patterns (shown by the gaps between the black signal) applied on different ECG waveforms. These are examples of the inputs used in the ECG Imputation and Heartbeat Detection Task. Here we can see that ECG missingness patterns are very complex in terms of their frequency and their duration. The ECG signals visualized here also are quite heterogeneous with many different morphologies, especially due to their potential differing lead of origin, with some signals having large peaks while others have large valleys, and different rhythms with some signals being more dense than others.}
\label{fig:bfmissingness}
\end{figure}

\begin{figure}[!htbp]
\centering
\includegraphics[width=.65\textwidth]{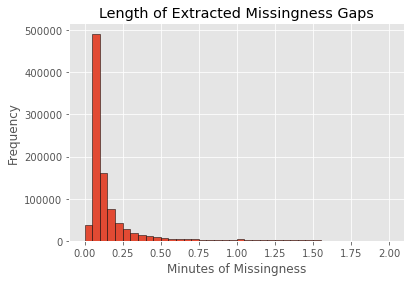}
\caption{Histogram of Length of Missingness Gaps found in Extracted ECG mHealth Missingness Patterns. The missingness gaps have a wide range: the majority of missingness gaps are 3-9 seconds long but some gaps can last more than a minute.}
\label{fig:bfmissdist}
\end{figure}

\pagebreak

\subsection{PPG Extracted mHealth Missingness} 

\begin{figure}[!htbp]
\centering
\includegraphics[width=\textwidth]{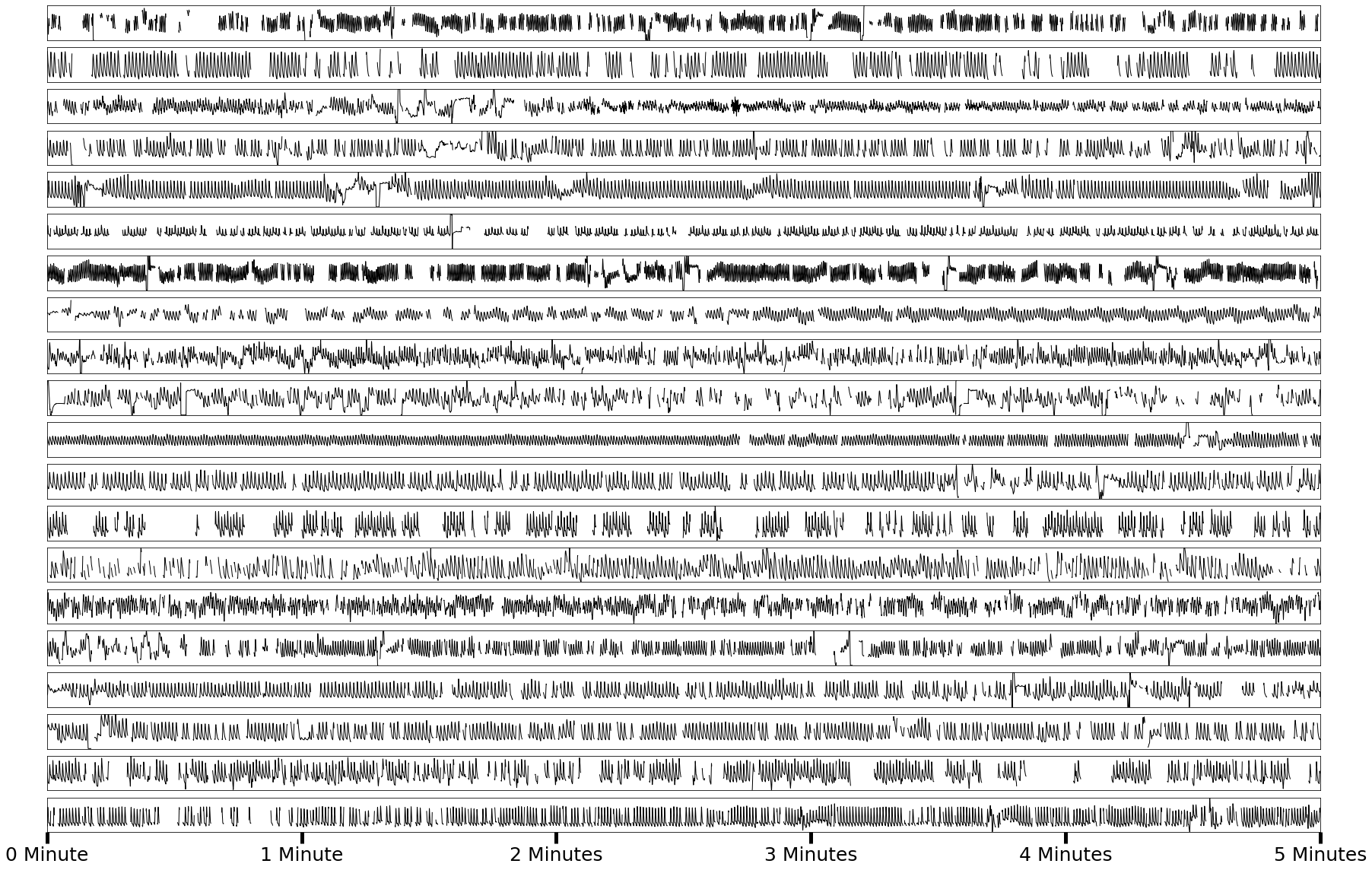}
\caption{ Various extracted PPG mHealth signal missingness patterns (shown by the gaps between the black signal) applied on different PPG waveforms. These are examples of the inputs used in the PPG Imputation and Heartbeat Detection Task. The PPG missingness patterns are different from those found in ECG, with much shorter gaps comparatively. The PPG signals are generally of simpler shapes, and there is more noise found in these PPG signals compared to the ECG signals.}
\label{fig:ppgbfmissingness}
\end{figure}

\begin{figure}[!htbp]
\centering
\includegraphics[width=.65\textwidth]{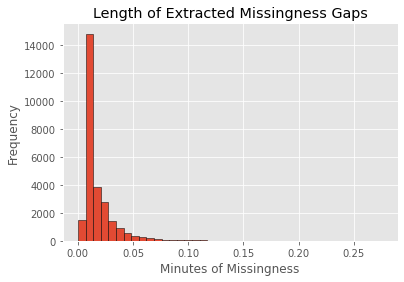}
\caption{Histogram of Length of Missingness Gaps found in Extracted PPG mHealth Missingness Patterns. As we visually saw in Figure A3, the missingness gaps in PPG mHealth signals are shorter.}
\label{fig:ppgbfmissdist}
\end{figure}


\newpage

\FloatBarrier
\newpage 
\section{Experimental Setup Details}
Models were trained on Titan Xp GPUs for 24 hours each or until convergence, whichever came first, on an internal Georgia Tech GPU Cluster. For each model trained on the 10-second-long ECG data used in the extended loss scenario for the ECG Imputation and Cardiac Classification Task, they are used to initialize the model for the ECG Imputation and Heartbeat Detection task before being further fine-tuned. Our PulseImpute repo (\textcolor{cyan}{\url{github.com/rehg-lab/pulseimpute}}) contains everything needed to reproduce results including a script to download the data and model checkpoints saved.

BRITS and NAOMI + BRITS w/ GAIL were implemented with their original paper's code base found \url{https://github.com/caow13/BRITS} and \url{https://github.com/felixykliu/NAOMI}, respectively. The training procedures were set up to be identical to the original, with the only modification being how missingness was simulated during training. Rather than their default missingness procedure of dropping out individual time-points independently and at random, they were trained on task-specific missingness patterns, as described in Sections 3.1, 3.2, 3.3.

For the transformer models, the longformer's dilated sliding window attention was used for the 5-minute-long data in Heartbeat detection. Conv9 uses the maximum kernel size for conv self-attention in its prior work \cite{convattn}, and our BDC module's query/key transformations have receptive fields of 883 ($\sim$9 sec). Each of the transformer-based architectures used follow the simple architecture scheme of one 1D Convolution Layer for embedding, two Transformer Encoder Layers, followed by one 1D Convolution Layer for projection for imputation.  The transformer models were trained with a Masked Predictive Coding procedure, introduced in \cite{jiang2019mpc}, inspired from the original Masked Language Prediction procedure, introduced in \cite{bert}. Given a block in which missingness would like to be ablated, there is a 80\% probability that it is replaced with a 0 vector, 10\% probability that sinusoidal vector is added as noise, and 10\% probability that the block is kept the same. L2 Loss is then calculated between the imputed result and ground-truth.

Please see our code repo for further details on hyperparameters, experimental set-up, and reproducibility.

\newpage 
\section{Extra Results and Visualizations} \label{sec:mimic_ecg_extraviz}
In this section, we show extra visualizations of the performance of each of the imputation models, grouped by their downstream task: ECG Imputation and Heartbeat Detection, PPG Imputation and Heartbeat Detection, and ECG Imputation and Cardiac Pathophysiology Classification. ECG/PPG Imputation and Heartbeat Detection Tasks benchmark imputation by applying extracted mHealth missingness patterns on 5-minute-long ECG/PPG data. ECG Imputation and Cardiac Classification benchmarks imputation by systematically varying amount of missingness with the extended and transient loss missingness models on 10-second-long ECG data. The purpose of this section is to visually evaluate the reconstruction quality of each of the models, as well as understanding the variation of the imputation model performance with density plots of MSE for each waveform.

\subsection{ECG Imputation and Heartbeat Detection}

\begin{figure}[!htbp]
\begin{center}
\includegraphics[width=1\textwidth]{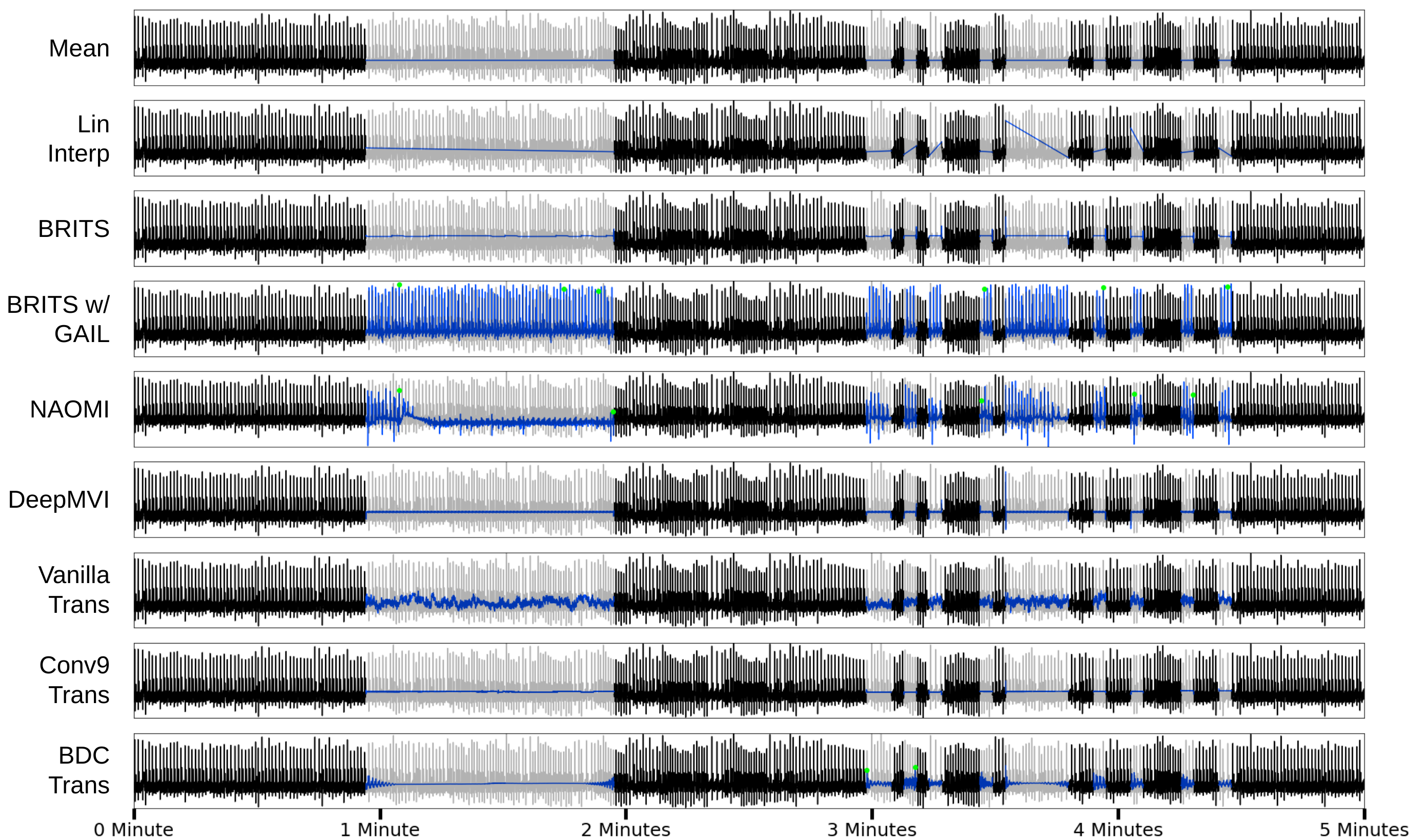}
\caption{ Extra visualization \#1 of 5 minutes of imputation results from ECG Heartbeat Detection. The green dots designate True Positive reconstructed peak detection. Grey designate the ground truth, blue the imputation results, and black the not-missing data. Here we see none of the models are able to perform well with long gaps of missingness. BRITS w/ GAIL appears to do well, but the small amount of green dots signifies that the rhythm of the original signal was unable to be reconstructed.}
\label{fig:mimicviz1}
\end{center}
\end{figure}

\begin{figure}[!htbp]
\begin{center}
\includegraphics[width=1\textwidth]{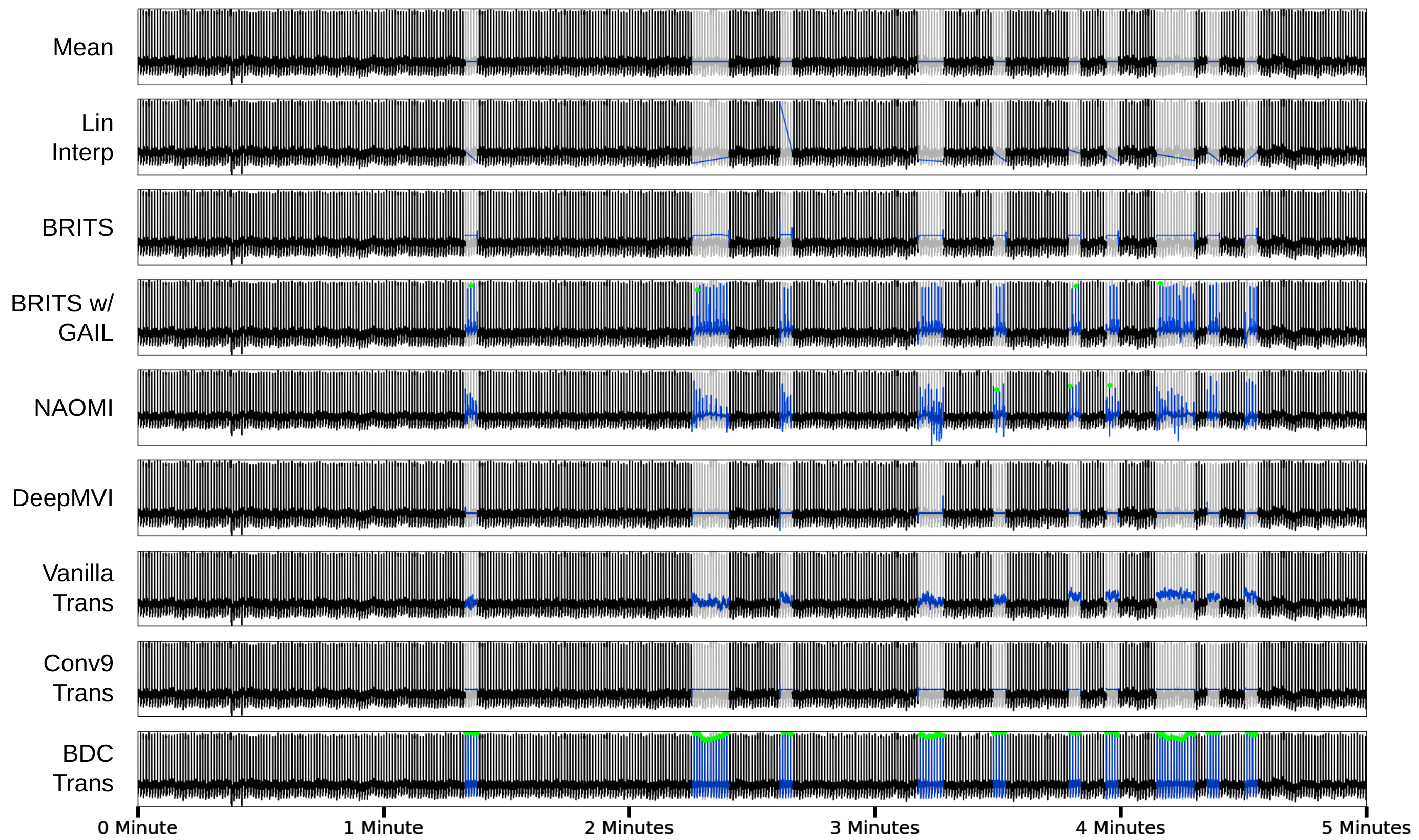}
\caption{ Extra visualization \#2 of 5 minutes of imputation results from ECG Heartbeat Detection. The green dots designate True Positive reconstructed peak detection. Grey designate the ground truth, blue the imputation results, and black the not-missing data. Given a signal with shorter missingness gaps, BDC transformer is able to reconstruct the signal and rhythm well, shown by the large amount of green dots.}
\label{fig:mimicviz2}
\end{center}
\end{figure}

\begin{figure}[!htbp]
\begin{center}
\includegraphics[width=1\textwidth]{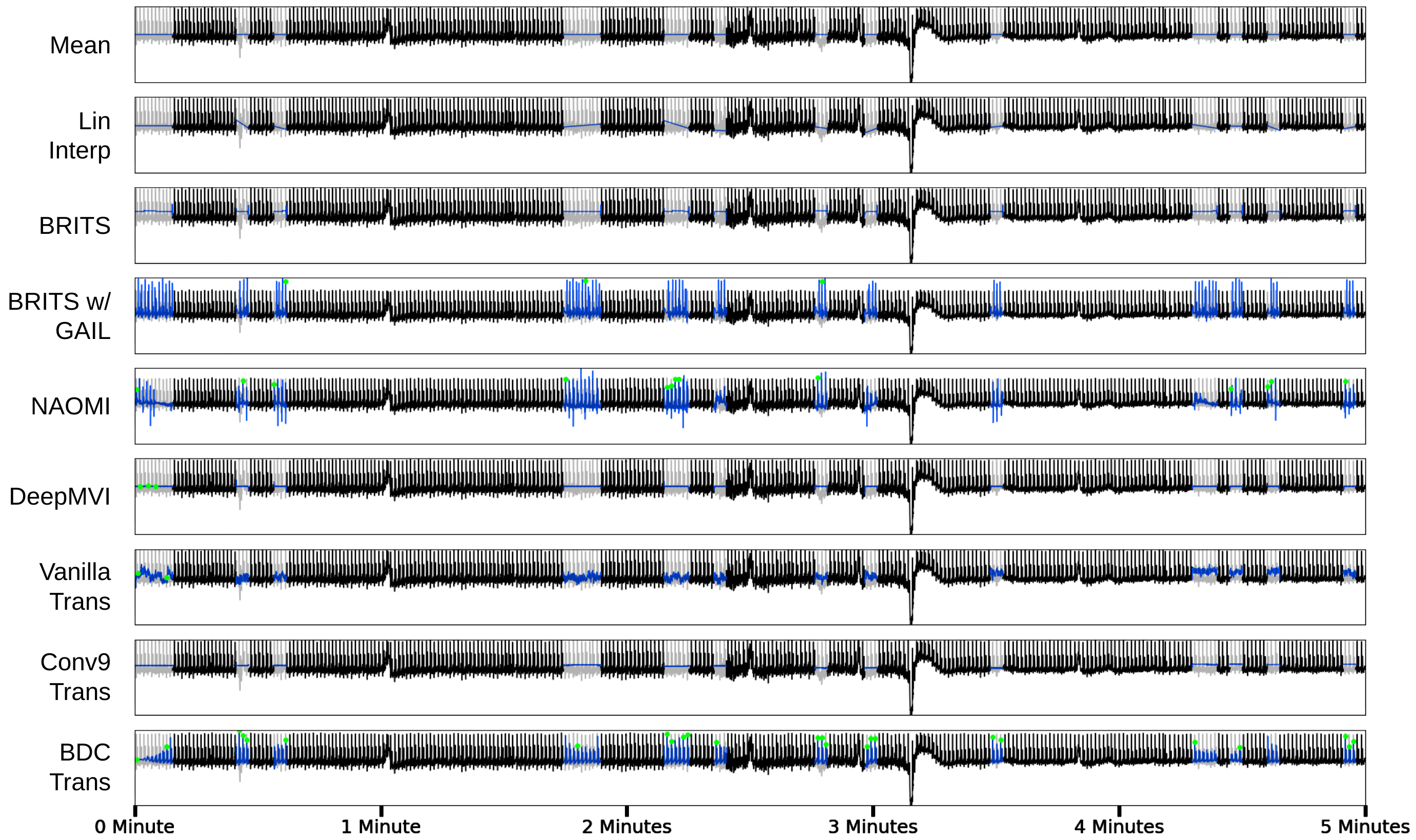}
\caption{ Extra visualization \#3 of 5 minutes of imputation results from ECG Heartbeat Detection. The green dots designate True Positive reconstructed peak detection. Grey designate the ground truth, blue the imputation results, and black the not-missing data.}
\label{fig:mimicviz3}
\end{center}
\end{figure}

\begin{figure}[!htbp]
\begin{center}
\includegraphics[width=.6\textwidth]{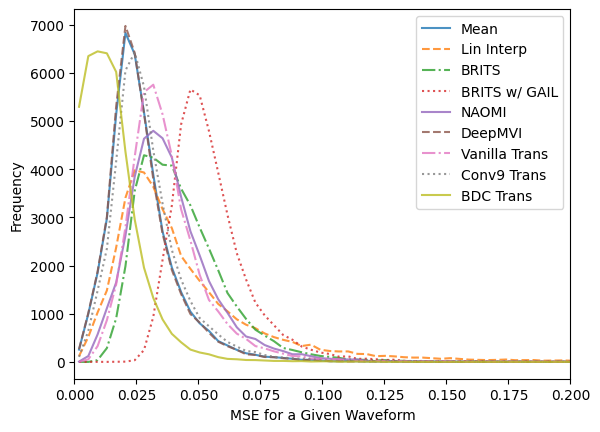}
\caption{ Density of MSE values for a given ECG waveform, for each imputation model. This demonstrates the variation of performance across all imputation methods, which shows that all existing imputation models have poor performance, with many models unable to achieve better MSE distributions than mean and linear interpolation. }
\label{fig:ecgmsedensity}
\end{center}
\end{figure}

\FloatBarrier
\subsection{PPG Imputation and Heartbeat Detection}

\begin{figure}[!htbp]
\begin{center}
\includegraphics[width=1\textwidth]{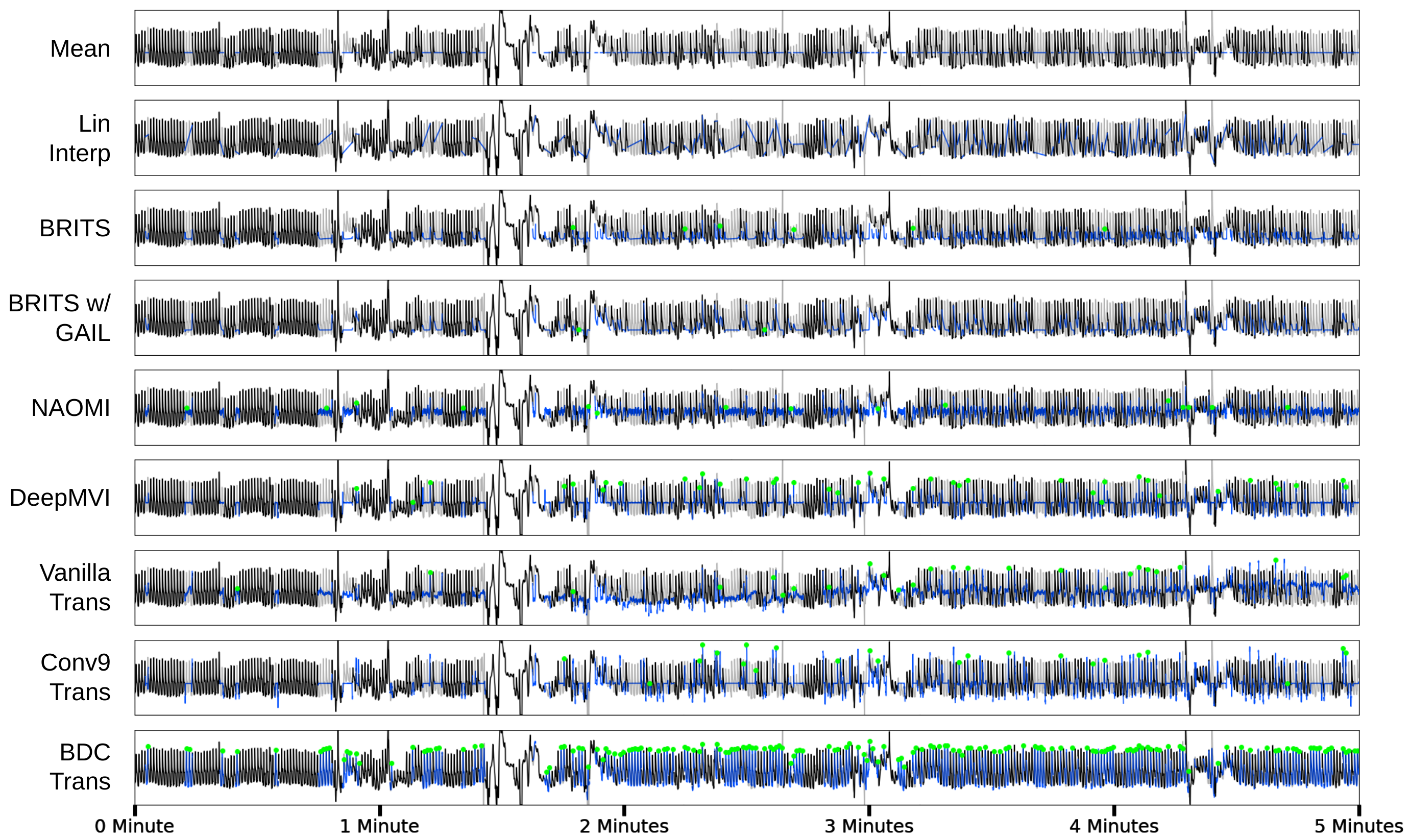}
\caption{ Extra visualization \#1 of 5 minutes of imputation results from PPG Heartbeat Detection. The green dots designate True Positive reconstructed peak detection. Grey designate the ground truth, blue the imputation results, and black the not-missing data. We can see that imputation models perform better in this PPG setting, with the simpler morphologies and shorter gaps of missingness. This can be seen with an increased amount of green dots for the ML models, which signify correct peak reconstruction. In general, BDC transformer has very strong performance, able to reconstruct the signal nearly perfectly.}
\label{fig:ppgmimicviz1}
\end{center}
\end{figure}

\begin{figure}[!htbp]
\begin{center}
\includegraphics[width=1\textwidth]{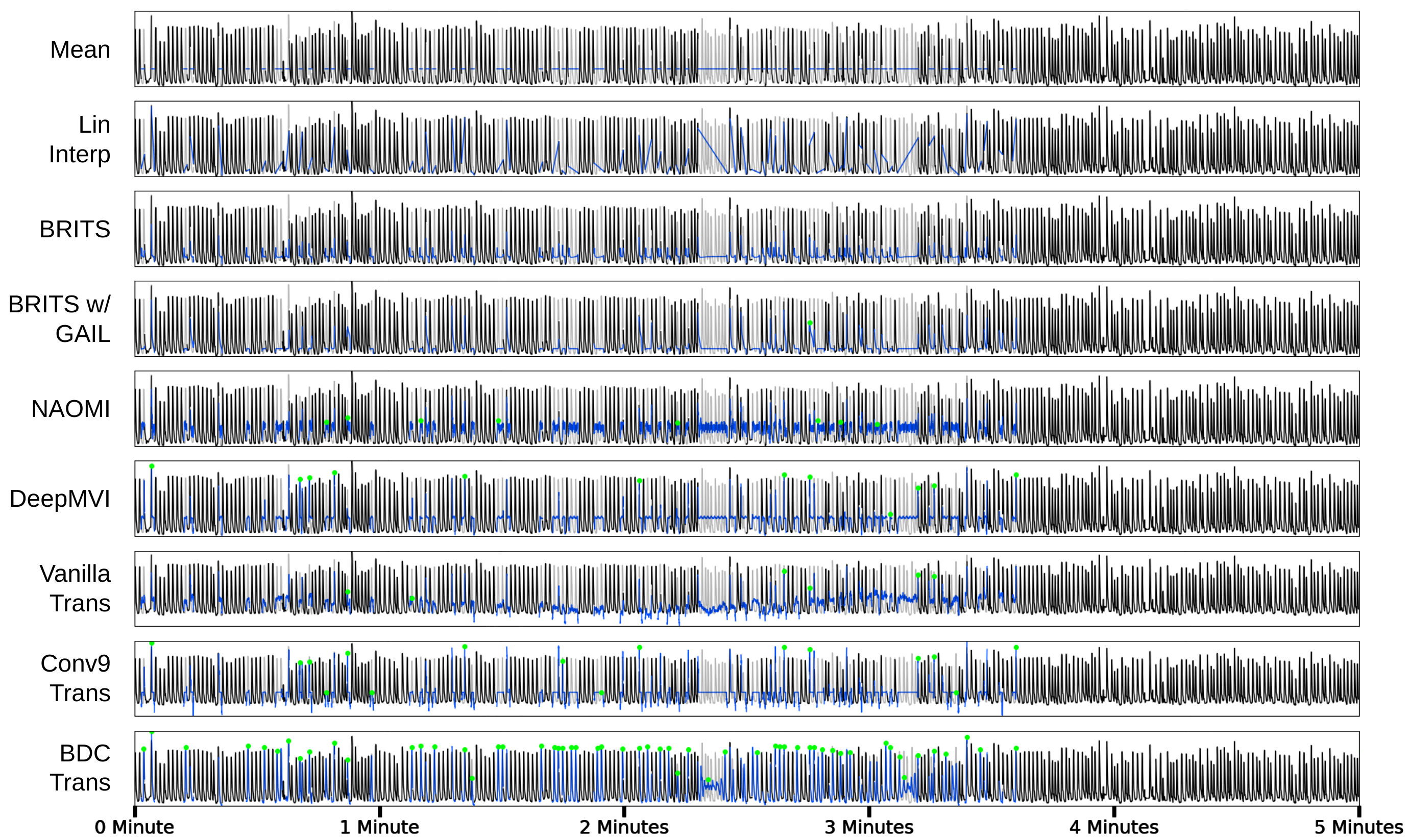}
\caption{ Extra visualization \#2 of 5 minutes of imputation results from PPG Heartbeat Detection. The green dots designate True Positive reconstructed peak detection. Grey designate the ground truth, blue the imputation results, and black the not-missing data.}
\label{fig:ppgmimicviz2}
\end{center}
\end{figure}

\begin{figure}[!htbp]
\begin{center}
\includegraphics[width=1\textwidth]{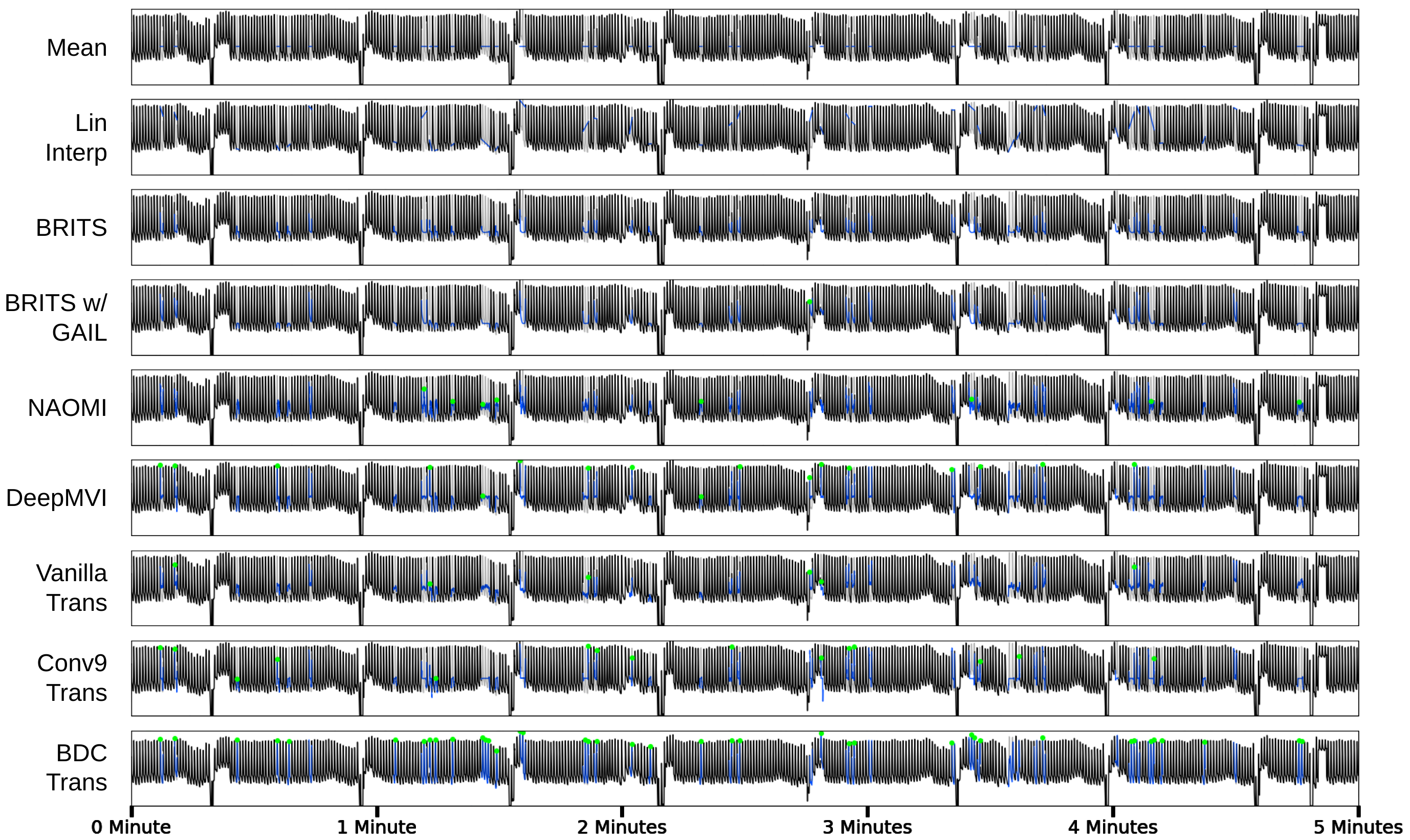}
\caption{ Extra visualization \#3 of 5 minutes of imputation results from PPG Heartbeat Detection. The green dots designate True Positive reconstructed peak detection. Grey designate the ground truth, blue the imputation results, and black the not-missing data.}
\label{fig:ppgmimicviz3}
\end{center}
\end{figure}

\begin{figure}[!htbp]
\begin{center}
\includegraphics[width=.6\textwidth]{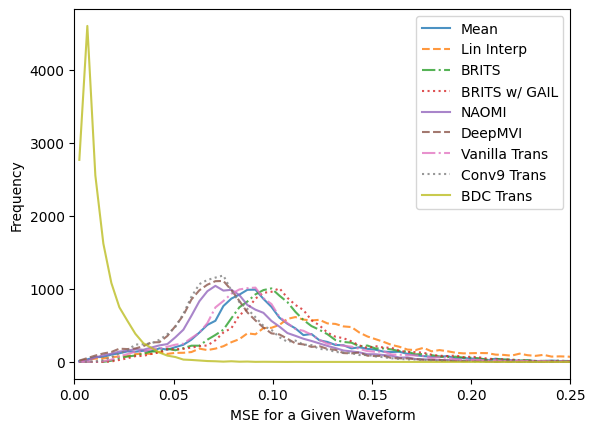}
\caption{ Density of MSE values for a given PPG waveform, for each imputation model. This demonstrates the variation of performance, which shows that compared to the ECG task, there are only a few models with better MSE distributions than the mean imputation model, namely Conv9 Transformer, DeepMVI, NAOMI, and our BDC transformer. This demonstrates that in general, this PPG imputation task is easier for the ML models than the ECG imputation counterpart, likely due to the simpler morphologies and shorter missingness gaps present.}
\label{fig:ppgmsedensity}
\end{center}
\end{figure}

\FloatBarrier
\newpage 
\subsection{ECG Imputation and Cardiac Pathophysiology Classification}
\begin{figure}[!htbp]
\begin{center}
\includegraphics[width=1\textwidth]{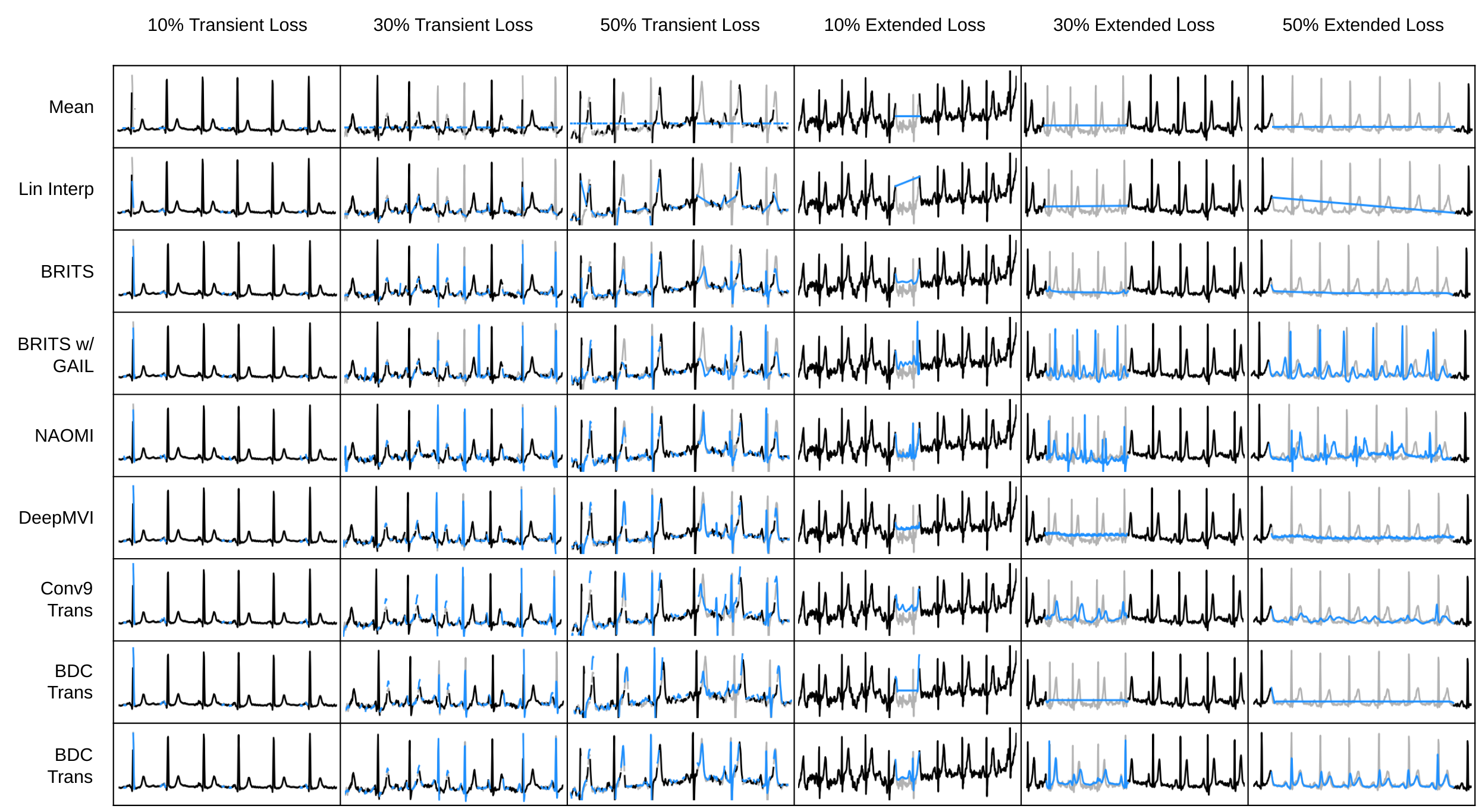}
\caption{ Extra visualization of imputation results from ECG Cardiac Classification, where each plot corresponds to a 6-second subsequence of the 10 second signal. Grey designate the ground truth, blue the imputation results, and black the not-missing data. In the transient setting, all models perform generally well. However, in the extended setting, we see the GAN methods (e.g. NAOMI and BRITS w/ GAIL) will reconstruct signals that do not mimic the non-ablated data, and the other methods (e.g. BRITS and DeepMVI) have flat-line imputations. BDC transformer matches the rhythm of the ablated signal well, but struggles with reproducing the R peaks (R peaks are the tallest steepest peaks in the signal). 
}
\label{fig:ptbxlextraviz}
\end{center}
\end{figure}

\begin{figure}[!htbp]
\begin{center}
\includegraphics[width=\textwidth]{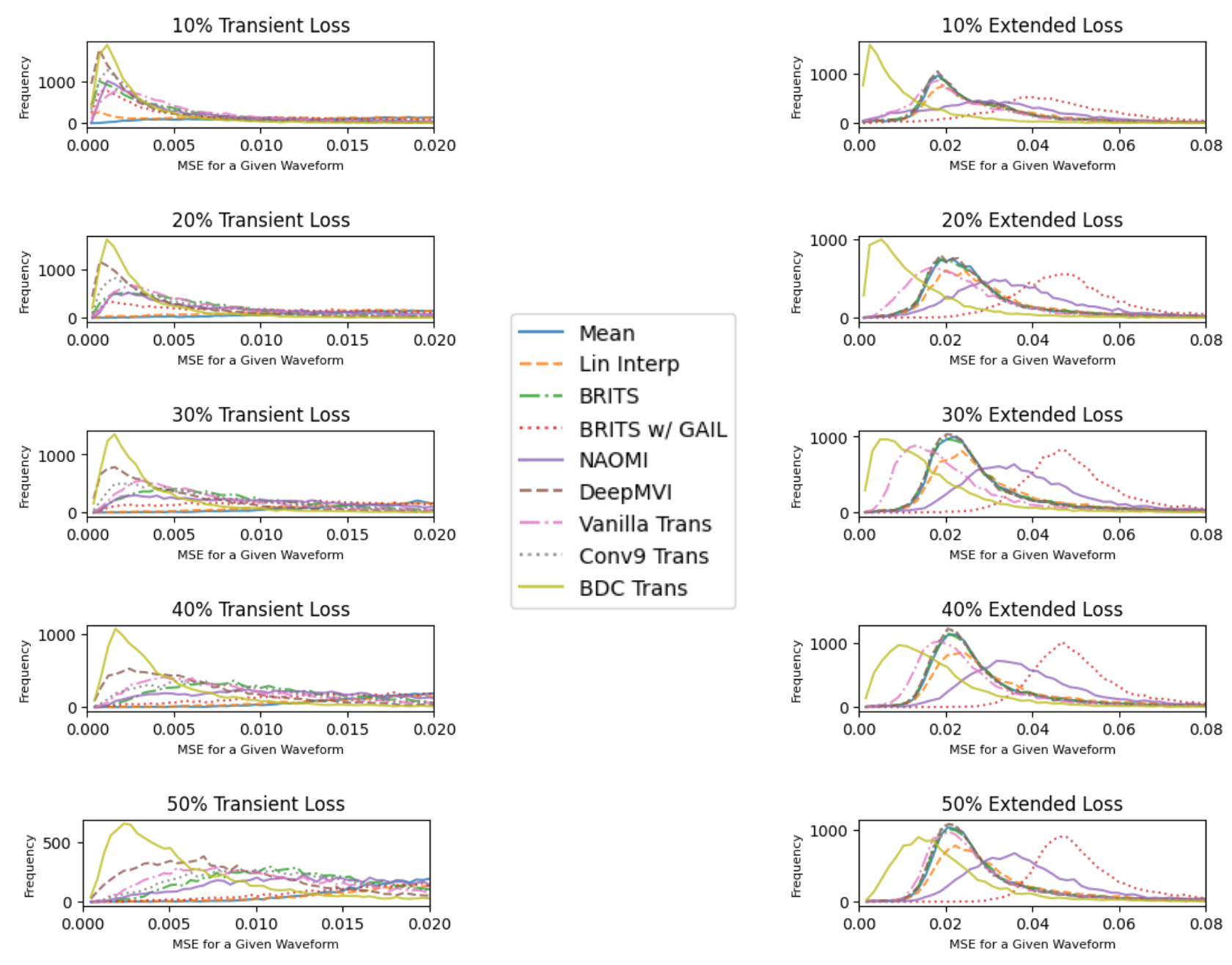}
\caption{ Density of MSE values for a given ECG waveform, for each imputation model for both transient and extended missingness scenarios. These plots demonstrate imputation models have a much easier time modeling in the transient loss scenario, with most models performing well. However, as missingness duration increases in transient loss, many models exhibit decreased performance, with an increased MSE. In extended loss, most models' performance stays constant (except for BDC and Vanilla transformer), imputations seemingly independent of amount of missingness. This makes sense if we also look at the visualizations in Figure \ref{fig:ptbxlextraviz} above. The GAN methods (e.g. NAOMI and BRITS w/ GAIL) do not seem to depend heavily on the non-ablated data because their reconstructions do not mimic the non-ablated data, and the other methods (e.g. BRITS and DeepMVI) have simple flat-line imputations, regardless of the amount missing.
 }
\label{fig:ptbxlecgmsedensity}
\end{center}
\end{figure}

\begin{table}[!htbp]
\centering
\resizebox{\textwidth}{!}{%
\begin{tabular}{@{}cccccccccc@{}}
\toprule
\multicolumn{1}{l}{} & \multicolumn{1}{l}{} & \multicolumn{4}{c}{\textbf{Transient}} & \multicolumn{4}{c}{\textbf{Extended}} \\ \cmidrule(l){3-6}  \cmidrule(l){7-10}
\multicolumn{1}{l}{\textbf{\% Miss}} & \textbf{Models} & \multicolumn{1}{c}{\textbf{MSE}} & \multicolumn{1}{c}{\textbf{Rhy AUC}} & \multicolumn{1}{c}{\textbf{Form AUC}} & \multicolumn{1}{c}{\textbf{Diag AUC}} & \multicolumn{1}{c}{\textbf{MSE}} & \multicolumn{1}{c}{\textbf{Rhy AUC}} & \multicolumn{1}{c}{\textbf{Form AUC}} & \multicolumn{1}{c}{\textbf{Diag AUC}} \\ \midrule
0 & - & \multicolumn{1}{c}{0} & \multicolumn{1}{c}{.949} & \multicolumn{1}{c}{.796} & \multicolumn{1}{c}{.845} & \multicolumn{1}{c}{0} & \multicolumn{1}{c}{.949} & \multicolumn{1}{c}{.796} & \multicolumn{1}{c}{.845} \\ \midrule
\multirow{14}{*}{10} & Mean & \multicolumn{1}{c}{.0302 ± .00044} & \multicolumn{1}{c}{.922 ± .0112} & \multicolumn{1}{l}{.776 ± .0186} & .827 ± .0120 & \multicolumn{1}{l}{.0300 ± .00042} & \multicolumn{1}{l}{.886 ± .0111} & \multicolumn{1}{l}{.786 ± .0166} & .827 ± .0199 \\ \cmidrule(l){2-10} 
 & Lin Interp & \multicolumn{1}{c}{.0221 ± .00033} & \multicolumn{1}{c}{.927 ± .0108} & \multicolumn{1}{l}{.788 ± .0187} & .833 ± .0117 & \multicolumn{1}{l}{.0438 ± .00107} & \multicolumn{1}{l}{.890 ± .0114} & \multicolumn{1}{l}{.787 ± .0156} & .824 ± .0190 \\ \cmidrule(l){2-10} 
 & FFT & \multicolumn{1}{c}{.0482 ± .00058} & \multicolumn{1}{c}{.923 ± .0125} & \multicolumn{1}{l}{.786 ± .0169} & .832 ± .0102 & \multicolumn{1}{l}{.0368 ± .00047} & \multicolumn{1}{l}{.905 ± .0005} & \multicolumn{1}{l}{.778 ± .0154} & .826 ± .0115 \\ \cmidrule(l){2-10} 
 & BRITS & \multicolumn{1}{c}{.0059 ± .00013} & \multicolumn{1}{c}{.946 ± .0104} & \multicolumn{1}{l}{.793 ± .0184} & .846 ± .0082 & \multicolumn{1}{l}{.0285 ± .00041} & \multicolumn{1}{l}{.887 ± .0106} & \multicolumn{1}{l}{.787 ± .0194} & .834 ± .0192 \\ \cmidrule(l){2-10} 
 & BRITS w/ GAIL & \multicolumn{1}{c}{.0110 ± .00024} & \multicolumn{1}{c}{.939 ± .0101} & \multicolumn{1}{l}{.793 ± .0181} & .846 ± .0094 & \multicolumn{1}{l}{.0486 ± .00049} & \multicolumn{1}{l}{.904 ± .0108} & \multicolumn{1}{l}{.787 ± .0166} & .831 ± .0136 \\ \cmidrule(l){2-10} 
 & NAOMI & \multicolumn{1}{c}{.0306 ± .00019} & \multicolumn{1}{c}{.946 ± .0103} & \multicolumn{1}{l}{.793 ± .0180} & .847 ± .0080 & \multicolumn{1}{l}{.0341 ± .00043} & \multicolumn{1}{l}{.929 ± .0096} & \multicolumn{1}{l}{.782 ± .0205} & .843 ± .0100 \\ \cmidrule(l){2-10} 
 & DeepMVI & \multicolumn{1}{c}{.0039 ± .00010} & \multicolumn{1}{c}{.947 ± .0104} & \multicolumn{1}{l}{.794 ± .0185} & .847 ± .0080 & \multicolumn{1}{l}{.0285 ± .00038} & \multicolumn{1}{l}{.875 ± .0110} & \multicolumn{1}{l}{.783 ± .0197} & .824 ± .0213 \\ \cmidrule(l){2-10} 
 & Van Trans & \multicolumn{1}{c}{.0065 ± .00014} & \multicolumn{1}{c}{.946 ± .0105} & \multicolumn{1}{l}{.791 ± .0183} & .846 ± .0081 & \multicolumn{1}{l}{.0260 ± .00039} & \multicolumn{1}{l}{.895 ± .0105} & \multicolumn{1}{l}{.787 ± .0180} & .831 ± .0164 \\ \cmidrule(l){2-10} 
 & Conv9 Trans & \multicolumn{1}{c}{.0049 ± .00012} & \multicolumn{1}{c}{.944 ± .0104} & \multicolumn{1}{l}{.791 ± .0185} & .847 ± .0083 & \multicolumn{1}{l}{.0288 ± .00041} & \multicolumn{1}{l}{.887 ± .0109} & \multicolumn{1}{l}{.787 ± .0182} & .830 ± .0203 \\ \cmidrule(l){2-10} 
 & BDC Trans & \multicolumn{1}{c}{.0030 ± .00007} & \multicolumn{1}{c}{.948 ± .0104} & \multicolumn{1}{l}{.795 ± .0185} & .847 ± .0078 & \multicolumn{1}{l}{.0116 ± .00027} & \multicolumn{1}{l}{.944 ± .0188} & \multicolumn{1}{l}{.790 ± .0086} & .844 ± .0086 \\ \midrule
\multirow{14}{*}{20} & Mean & \multicolumn{1}{c}{.0302 ± .00038} & \multicolumn{1}{c}{.876 ± .0118} & \multicolumn{1}{l}{.762 ± .0194} & .805 ± .0217 & \multicolumn{1}{l}{.0301 ± .00039} & \multicolumn{1}{l}{.882 ± .0206} & \multicolumn{1}{l}{.775 ± .0230} & .811 ± .0230 \\ \cmidrule(l){2-10} 
 & Lin Interp & \multicolumn{1}{c}{.0239 ± .00028} & \multicolumn{1}{c}{.910 ± .0114} & \multicolumn{1}{l}{.782 ± .0196} & .813 ± .0134 & \multicolumn{1}{l}{.0454 ± .00104} & \multicolumn{1}{l}{.886 ± .0203} & \multicolumn{1}{l}{.774 ± .0215} & .812 ± .0215 \\ \cmidrule(l){2-10} 
 & FFT & \multicolumn{1}{c}{.0477 ± .00049} & \multicolumn{1}{c}{.890 ± .0147} & \multicolumn{1}{l}{.765 ± .0177} & .808 ± .0108 & \multicolumn{1}{l}{.0357 ± .00045} & \multicolumn{1}{l}{.871 ± .0226} & \multicolumn{1}{l}{.758 ± .0173} & .799 ± .0126 \\ \cmidrule(l){2-10} 
 & BRITS & \multicolumn{1}{c}{.0075 ± .00013} & \multicolumn{1}{c}{.944 ± .0104} & \multicolumn{1}{l}{.793 ± .0160} & .844 ± .0076 & \multicolumn{1}{l}{.0296 ± .00040} & \multicolumn{1}{l}{.883 ± .0113} & \multicolumn{1}{l}{.776 ± .0206} & .825 ± .0242 \\ \cmidrule(l){2-10} 
 & BRITS w/ GAIL & \multicolumn{1}{c}{.0137 ± .00023} & \multicolumn{1}{c}{.929 ± .0100} & \multicolumn{1}{l}{.787 ± .0179} & .838 ± .0095 & \multicolumn{1}{l}{.0523 ± .00045} & \multicolumn{1}{l}{.898 ± .0109} & \multicolumn{1}{l}{.773 ± .0164} & .816 ± .0145 \\ \cmidrule(l){2-10} 
 & NAOMI & \multicolumn{1}{c}{.0102 ± .00018} & \multicolumn{1}{c}{.941 ± .0099} & \multicolumn{1}{l}{.792 ± .0158} & .846 ± .0088 & \multicolumn{1}{l}{.0388 ± .00039} & \multicolumn{1}{l}{.903 ± .0113} & \multicolumn{1}{l}{.771 ± .0189} & .827 ± .0136 \\ \cmidrule(l){2-10} 
 & DeepMVI & \multicolumn{1}{c}{.0045 ± .00010} & \multicolumn{1}{c}{.945 ± .0107} & \multicolumn{1}{l}{.794 ± .0172} & .845 ± .0076 & \multicolumn{1}{l}{.0288 ± .00035} & \multicolumn{1}{l}{.845 ± .0136} & \multicolumn{1}{l}{.771 ± .0212} & .803 ± .0279 \\ \cmidrule(l){2-10} 
 & Van Trans & \multicolumn{1}{c}{.0071 ± .00014} & \multicolumn{1}{c}{.944 ± .0104} & \multicolumn{1}{l}{.790 ± .0164} & .845 ± .0081 & \multicolumn{1}{l}{.0251 ± .00036} & \multicolumn{1}{l}{.895 ± .0122} & \multicolumn{1}{l}{.774 ± .0191} & .808 ± .0189 \\ \cmidrule(l){2-10} 
 & Conv9 Trans & \multicolumn{1}{c}{.0060 ± .00012} & \multicolumn{1}{c}{.938 ± .0105} & \multicolumn{1}{l}{.789 ± .0181} & .842 ± .0088 & \multicolumn{1}{l}{.0293 ± .00039} & \multicolumn{1}{l}{.882 ± .0120} & \multicolumn{1}{l}{.777 ± .0204} & .819 ± .0232 \\ \cmidrule(l){2-10} 
 & BDC Trans & \multicolumn{1}{c}{.0033 ± .00007} & \multicolumn{1}{c}{.947 ± .0105} & \multicolumn{1}{l}{.795 ± .0160} & .847 ± .0076 & \multicolumn{1}{l}{.0136 ± .00027} & \multicolumn{1}{l}{.935 ± .0119} & \multicolumn{1}{l}{.780 ± .0196} & .829 ± .0093 \\ \midrule
\multirow{14}{*}{30} & Mean & \multicolumn{1}{c}{.0302 ± .00036} & \multicolumn{1}{c}{.828 ± .0119} & \multicolumn{1}{l}{.735 ± .0212} & .782 ± .0240 & \multicolumn{1}{l}{.0304 ± .00040} & \multicolumn{1}{l}{.883 ± .0129} & \multicolumn{1}{l}{.760 ± .0197} & .807 ± .0204 \\ \cmidrule(l){2-10} 
 & Lin Interp & \multicolumn{1}{c}{.0260 ± .00025} & \multicolumn{1}{c}{.870 ± .0119} & \multicolumn{1}{l}{.766 ± .0181} & .792 ± .0207 & \multicolumn{1}{l}{.0454 ± .00109} & \multicolumn{1}{l}{.879 ± .0123} & \multicolumn{1}{l}{.761 ± .0190} & .801 ± .0204 \\ \cmidrule(l){2-10} 
 & FFT & \multicolumn{1}{c}{.0470 ± .00046} & \multicolumn{1}{c}{.842 ± .0236} & \multicolumn{1}{l}{.738 ± .0198} & .780 ± .0114 & \multicolumn{1}{l}{.0344 ± .00043} & \multicolumn{1}{l}{.870 ± .0182} & \multicolumn{1}{l}{.746 ± .0161} & .768 ± .0137 \\ \cmidrule(l){2-10} 
 & BRITS & \multicolumn{1}{c}{.0095 ± .00014} & \multicolumn{1}{c}{.933 ± .0099} & \multicolumn{1}{l}{.783 ± .0179} & .838 ± .0179 & \multicolumn{1}{l}{.0299 ± .00039} & \multicolumn{1}{l}{.880 ± .0115} & \multicolumn{1}{l}{.766 ± .0202} & .823 ± .0188 \\ \cmidrule(l){2-10} 
 & BRITS w/ GAIL & \multicolumn{1}{c}{.0173 ± .00023} & \multicolumn{1}{c}{.909 ± .0101} & \multicolumn{1}{l}{.777 ± .0192} & .826 ± .0192 & \multicolumn{1}{l}{.0535 ± .00043} & \multicolumn{1}{l}{.890 ± .0107} & \multicolumn{1}{l}{.768 ± .0180} & .815 ± .0133 \\ \cmidrule(l){2-10} 
 & NAOMI & \multicolumn{1}{c}{.0124 ± .00018} & \multicolumn{1}{c}{.932 ± .0099} & \multicolumn{1}{l}{.781 ± .0167} & .840 ± .0167 & \multicolumn{1}{l}{.0405 ± .00038} & \multicolumn{1}{l}{.899 ± .0119} & \multicolumn{1}{l}{.751 ± .0229} & .808 ± .0113 \\ \cmidrule(l){2-10} 
 & DeepMVI & \multicolumn{1}{c}{.0056 ± .00011} & \multicolumn{1}{c}{.939 ± .0108} & \multicolumn{1}{l}{.790 ± .0155} & .841 ± .0155 & \multicolumn{1}{l}{.0290 ± .00036} & \multicolumn{1}{l}{.856 ± .0136} & \multicolumn{1}{l}{.751 ± .0210} & .797 ± .0223 \\ \cmidrule(l){2-10} 
 & Van Trans & \multicolumn{1}{c}{.0084 ± .00014} & \multicolumn{1}{c}{.936 ± .0107} & \multicolumn{1}{l}{.786 ± .0152} & .841 ± .0152 & \multicolumn{1}{l}{.0226 ± .00035} & \multicolumn{1}{l}{.903 ± .0117} & \multicolumn{1}{l}{.758 ± .0194} & .796 ± .0143 \\ \cmidrule(l){2-10} 
 & Conv9 Trans & \multicolumn{1}{c}{.0078 ± .00013} & \multicolumn{1}{c}{.930 ± .0106} & \multicolumn{1}{l}{.783 ± .0168} & .837 ± .0168 & \multicolumn{1}{l}{.0294 ± .00038} & \multicolumn{1}{l}{.885 ± .0119} & \multicolumn{1}{l}{.761 ± .0207} & .814 ± .0190 \\ \cmidrule(l){2-10} 
 & BDC Trans & \multicolumn{1}{c}{.0038 ± .00007} & \multicolumn{1}{c}{.945 ± .0105} & \multicolumn{1}{l}{.793 ± .0177} & .844 ± .0177 & \multicolumn{1}{l}{.0159 ± .00028} & \multicolumn{1}{l}{.930 ± .0121} & \multicolumn{1}{l}{.773 ± .0197} & .817 ± .0094 \\ \midrule
\multirow{14}{*}{40} & Mean & \multicolumn{1}{c}{.0302 ± .00035} & \multicolumn{1}{c}{.784 ± .0128} & \multicolumn{1}{l}{.707 ± .0240} & .752 ± .0279 & \multicolumn{1}{l}{.0307 ± .00039} & \multicolumn{1}{l}{.871 ± .0114} & \multicolumn{1}{l}{.761 ± .0172} & .812 ± .0181 \\ \cmidrule(l){2-10} 
 & Lin Interp & \multicolumn{1}{c}{.0282 ± .00026} & \multicolumn{1}{c}{.827 ± .0130} & \multicolumn{1}{l}{.745 ± .0180} & .766 ± .0275 & \multicolumn{1}{l}{.0460 ± .00106} & \multicolumn{1}{l}{.870 ± .0110} & \multicolumn{1}{l}{.750 ± .0188} & .804 ± .0154 \\ \cmidrule(l){2-10} 
 & FFT & \multicolumn{1}{c}{.0461 ± .00047} & \multicolumn{1}{c}{.776 ± .0298} & \multicolumn{1}{l}{.698 ± .0235} & .740 ± .0143 & \multicolumn{1}{l}{.0332 ± .00043} & \multicolumn{1}{l}{.843 ± .0176} & \multicolumn{1}{l}{.736 ± .0183} & .762 ± .0134 \\ \cmidrule(l){2-10} 
 & BRITS & \multicolumn{1}{c}{.0121 ± .00016} & \multicolumn{1}{c}{.923 ± .0109} & \multicolumn{1}{l}{.778 ± .0161} & .832 ± .0098 & \multicolumn{1}{l}{.0301 ± .00038} & \multicolumn{1}{l}{.870 ± .0107} & \multicolumn{1}{l}{.762 ± .0206} & .825 ± .0190 \\ \cmidrule(l){2-10} 
 & BRITS w/ GAIL & \multicolumn{1}{c}{.0216 ± .00024} & \multicolumn{1}{c}{.870 ± .0101} & \multicolumn{1}{l}{.764 ± .0177} & .807 ± .0182 & \multicolumn{1}{l}{.0543 ± .00041} & \multicolumn{1}{l}{.874 ± .0113} & \multicolumn{1}{l}{.754 ± .0184} & .809 ± .0147 \\ \cmidrule(l){2-10} 
 & NAOMI & \multicolumn{1}{c}{.0150 ± .00018} & \multicolumn{1}{c}{.922 ± .0099} & \multicolumn{1}{l}{.775 ± .0171} & .833 ± .0110 & \multicolumn{1}{l}{.0410 ± .00036} & \multicolumn{1}{l}{.876 ± .0122} & \multicolumn{1}{l}{.749 ± .0160} & .801 ± .0166 \\ \cmidrule(l){2-10} 
 & DeepMVI & \multicolumn{1}{c}{.0072 ± .00012} & \multicolumn{1}{c}{.925 ± .0110} & \multicolumn{1}{l}{.781 ± .0169} & .833 ± .0102 & \multicolumn{1}{l}{.0291 ± .00034} & \multicolumn{1}{l}{.830 ± .0127} & \multicolumn{1}{l}{.741 ± .0248} & .799 ± .0297 \\ \cmidrule(l){2-10} 
 & Van Trans & \multicolumn{1}{c}{.0105 ± .00016} & \multicolumn{1}{c}{.929 ± .0107} & \multicolumn{1}{l}{.774 ± .0164} & .835 ± .0103 & \multicolumn{1}{l}{.0270 ± .00036} & \multicolumn{1}{l}{.860 ± .0120} & \multicolumn{1}{l}{.755 ± .0182} & .790 ± .0219 \\ \cmidrule(l){2-10} 
 & Conv9 Trans & \multicolumn{1}{c}{.0106 ± .00016} & \multicolumn{1}{c}{.915 ± .0109} & \multicolumn{1}{l}{.768 ± .0190} & .827 ± .0122 & \multicolumn{1}{l}{.0295 ± .00037} & \multicolumn{1}{l}{.870 ± .0115} & \multicolumn{1}{l}{.760 ± .0182} & .815 ± .0175 \\ \cmidrule(l){2-10} 
 & BDC Trans & \multicolumn{1}{c}{.0048 ± .00009} & \multicolumn{1}{c}{.944 ± .0109} & \multicolumn{1}{l}{.790 ± .0183} & .841 ± .0084 & \multicolumn{1}{l}{.0181 ± .00029} & \multicolumn{1}{l}{.912 ± .0132} & \multicolumn{1}{l}{.762 ± .0194} & .801 ± .0113 \\ \midrule
\multirow{14}{*}{50} & Mean & \multicolumn{1}{c}{.0302 ± .00034} & \multicolumn{1}{c}{.758 ± .0137} & \multicolumn{1}{l}{.677 ± .0210} & .717 ± .0294 & \multicolumn{1}{l}{.0312 ± .00040} & \multicolumn{1}{l}{.858 ± .0119} & \multicolumn{1}{l}{.742 ± .0217} & .806 ± .0225 \\ \cmidrule(l){2-10} 
 & Lin Interp & \multicolumn{1}{c}{.0306 ± .00028} & \multicolumn{1}{c}{.772 ± .0130} & \multicolumn{1}{l}{.726 ± .0174} & .743 ± .0312 & \multicolumn{1}{l}{.0467 ± .00104} & \multicolumn{1}{l}{.846 ± .0113} & \multicolumn{1}{l}{.740 ± .0237} & .800 ± .0251 \\ \cmidrule(l){2-10} 
 & FFT & \multicolumn{1}{c}{.0448 ± .00046} & \multicolumn{1}{c}{.721 ± .0296} & \multicolumn{1}{l}{.665 ± .0216} & .706 ± .0138 & \multicolumn{1}{l}{.0325 ± .00038} & \multicolumn{1}{l}{.830 ± .0314} & \multicolumn{1}{l}{.727 ± .0219} & .766 ± .0129 \\ \cmidrule(l){2-10} 
 & BRITS & \multicolumn{1}{c}{.0155 ± .00018} & \multicolumn{1}{c}{.895 ± .0115} & \multicolumn{1}{l}{.761 ± .0151} & .817 ± .0142 & \multicolumn{1}{l}{.0302 ± .00036} & \multicolumn{1}{l}{.850 ± .0103} & \multicolumn{1}{l}{.744 ± .0212} & .825 ± .0242 \\ \cmidrule(l){2-10} 
 & BRITS w/ GAIL & \multicolumn{1}{c}{.0275 ± .00027} & \multicolumn{1}{c}{.841 ± .0117} & \multicolumn{1}{l}{.737 ± .0196} & .786 ± .0188 & \multicolumn{1}{l}{.0549 ± .00040} & \multicolumn{1}{l}{.860 ± .0120} & \multicolumn{1}{l}{.749 ± .0181} & .802 ± .0212 \\ \cmidrule(l){2-10} 
 & NAOMI & \multicolumn{1}{c}{.0179 ± .00019} & \multicolumn{1}{c}{.913 ± .0095} & \multicolumn{1}{l}{.765 ± .0155} & .829 ± .0101 & \multicolumn{1}{l}{.0411 ± .00037} & \multicolumn{1}{l}{.863 ± .0123} & \multicolumn{1}{l}{.716 ± .0201} & .786 ± .0167 \\ \cmidrule(l){2-10} 
 & DeepMVI & \multicolumn{1}{c}{.0098 ± .00014} & \multicolumn{1}{c}{.904 ± .0116} & \multicolumn{1}{l}{.767 ± .0170} & .820 ± .0136 & \multicolumn{1}{l}{.0294 ± .00034} & \multicolumn{1}{l}{.818 ± .0120} & \multicolumn{1}{l}{.736 ± .0222} & .797 ± .0225 \\ \cmidrule(l){2-10} 
 & Van Trans & \multicolumn{1}{c}{.0138 ± .00020} & \multicolumn{1}{c}{.903 ± .0108} & \multicolumn{1}{l}{.756 ± .0178} & .826 ± .0149 & \multicolumn{1}{l}{.0290 ± .00035} & \multicolumn{1}{l}{.827 ± .0133} & \multicolumn{1}{l}{.748 ± .0163} & .779 ± .0261 \\ \cmidrule(l){2-10} 
 & Conv9 Trans & \multicolumn{1}{c}{.0146 ± .00020} & \multicolumn{1}{c}{.877 ± .0117} & \multicolumn{1}{l}{.756 ± .0187} & .812 ± .0210 & \multicolumn{1}{l}{.0296 ± .00035} & \multicolumn{1}{l}{.856 ± .0111} & \multicolumn{1}{l}{.744 ± .0217} & .812 ± .0233 \\ \cmidrule(l){2-10} 
 & BDC Trans & \multicolumn{1}{c}{.0066 ± .00011} & \multicolumn{1}{c}{.936 ± .0109} & \multicolumn{1}{l}{.784 ± .0157} & .836 ± .0083 & \multicolumn{1}{l}{.0209 ± .00029} & \multicolumn{1}{l}{.892 ± .0126} & \multicolumn{1}{l}{.754 ± .0176} & .794 ± .0142 \\ \bottomrule
\end{tabular}%
}
\caption{Full tabulated results from the ECG cardiac classification task with Macro AUC values  with 95\% CI}
\label{tbl:ptbxl}
\end{table}

\begin{table}[]
\centering
\resizebox{\textwidth}{!}{%
\begin{tabular}{@{}cccccccccc@{}}
\toprule
\multicolumn{1}{l}{} & \multicolumn{1}{l}{} & \multicolumn{4}{c}{\textbf{Transient}} & \multicolumn{4}{c}{\textbf{Extended}} \\ \cmidrule(l){3-6}  \cmidrule(l){7-10}
\multicolumn{1}{l}{\textbf{\% Miss}} & \textbf{Models} & \multicolumn{1}{c}{\textbf{MSE}} & \multicolumn{1}{c}{\textbf{Rhy AUC}} & \multicolumn{1}{c}{\textbf{Form AUC}} & \multicolumn{1}{c}{\textbf{Diag AUC}} & \multicolumn{1}{c}{\textbf{MSE}} & \multicolumn{1}{c}{\textbf{Rhy AUC}} & \multicolumn{1}{c}{\textbf{Form AUC}} & \multicolumn{1}{c}{\textbf{Diag AUC}} \\ \midrule
0 &
  - &
  \multicolumn{1}{c}{0} &
  \multicolumn{1}{c}{.930} &
  \multicolumn{1}{c}{.752} &
  .794 &
  \multicolumn{1}{c}{0} &
  \multicolumn{1}{c}{.930} &
  \multicolumn{1}{c}{.752} &
  .794 \\ \midrule
\multirow{14}{*}{10} &
  Mean &
  \multicolumn{1}{c}{.0334 ± .00014} &
  \multicolumn{1}{c}{.904 ± .0051} &
  \multicolumn{1}{c}{.743 ± .0055} &
  .783 ± .0038 &
  \multicolumn{1}{c}{.0333 ± .00015} &
  \multicolumn{1}{c}{.885 ± .0053} &
  \multicolumn{1}{c}{.733 ± .0054} &
  .773 ± .0036 \\ \cmidrule(l){2-10} 
 &
  Lin Interp &
  \multicolumn{1}{c}{.0229 ± .00011} &
  \multicolumn{1}{c}{.919 ± .0045} &
  \multicolumn{1}{c}{.747 ± .0053} &
  .786 ± .0038 &
  \multicolumn{1}{c}{.0460 ± .00031} &
  \multicolumn{1}{c}{.880 ± .0056} &
  \multicolumn{1}{c}{.730 ± .0053} &
  .773 ± .0039 \\ \cmidrule(l){2-10} 
 &
  FFT &
  \multicolumn{1}{c}{.0487 ± .00017} &
  \multicolumn{1}{c}{.917 ± .0047} &
  \multicolumn{1}{c}{.744 ± .0054} &
  .774 ± .0036 &
  \multicolumn{1}{c}{.0406 ± .00017} &
  \multicolumn{1}{c}{.888 ± .0055} &
  \multicolumn{1}{c}{.734 ± .0051} &
  .783 ± .0037 \\ \cmidrule(l){2-10} 
 &
  BRITS &
  \multicolumn{1}{c}{.0036 ± .00003} &
  \multicolumn{1}{c}{.931 ± .0042} &
  \multicolumn{1}{c}{.749 ± .0052} &
  .795 ± .0037 &
  \multicolumn{1}{c}{.0336 ± .00016} &
  \multicolumn{1}{c}{.887 ± .0053} &
  \multicolumn{1}{c}{.741 ± .0052} &
  .777 ± .0038 \\ \cmidrule(l){2-10} 
 &
  BRITS w/ GAIL &
  \multicolumn{1}{c}{.0378 ± .00017} &
  \multicolumn{1}{c}{.887 ± .0056} &
  \multicolumn{1}{c}{.729 ± .0057} &
  .764 ± .0037 &
  \multicolumn{1}{c}{.4249 ± .00278} &
  \multicolumn{1}{c}{.883 ± .0051} &
  \multicolumn{1}{c}{.714 ± .0058} &
  .766 ± .0038 \\ \cmidrule(l){2-10} 
 &
  NAOMI &
  \multicolumn{1}{c}{.0264 ± .00015} &
  \multicolumn{1}{c}{.914 ± .0045} &
  \multicolumn{1}{c}{.743 ± .0054} &
  .783 ± .0038 &
  \multicolumn{1}{c}{.0409 ± .00021} &
  \multicolumn{1}{c}{.889 ± .0052} &
  \multicolumn{1}{c}{.731 ± .0057} &
  .776 ± .0038 \\ \cmidrule(l){2-10} 
 &
  DeepMVI &
  \multicolumn{1}{c}{1.1129 ± .00108} &
  \multicolumn{1}{c}{.609 ± .0085} &
  \multicolumn{1}{c}{.598 ± .0061} &
  .647 ± .0043 &
  \multicolumn{1}{c}{.0309 ± .00012} &
  \multicolumn{1}{c}{.878 ± .0055} &
  \multicolumn{1}{c}{.740 ± .0054} &
  .775 ± .0038 \\ \cmidrule(l){2-10} 
 &
  Van Trans &
  \multicolumn{1}{c}{.0015 ± .00002} &
  \multicolumn{1}{c}{.930 ± .0042} &
  \multicolumn{1}{c}{.750 ± .0051} &
  .794 ± .0037 &
  \multicolumn{1}{c}{.0304 ± .00012} &
  \multicolumn{1}{c}{.875 ± .0055} &
  \multicolumn{1}{c}{.735 ± .0053} &
  .764 ± .0038 \\ \cmidrule(l){2-10} 
 &
  Conv9 Trans &
  \multicolumn{1}{c}{.0018 ± .00002} &
  \multicolumn{1}{c}{.928 ± .0043} &
  \multicolumn{1}{c}{.750 ± .0051} &
  .793 ± .0037 &
  \multicolumn{1}{c}{.0323 ± .00015} &
  \multicolumn{1}{c}{.886 ± .0051} &
  \multicolumn{1}{c}{.734 ± .0054} &
  .776 ± .0037 \\ \cmidrule(l){2-10} 
 &
  BDC Trans &
  \multicolumn{1}{c}{.0015 ± .00001} &
  \multicolumn{1}{c}{.932 ± .0040} &
  \multicolumn{1}{c}{.749 ± .0054} &
  .794 ± .0037 &
  \multicolumn{1}{c}{.0132 ± .00010} &
  \multicolumn{1}{c}{.867 ± .0061} &
  \multicolumn{1}{c}{.713 ± .0053} &
  .751 ± .0042 \\ \midrule
\multirow{14}{*}{20} &
  Mean &
  \multicolumn{1}{c}{.0334 ± .00012} &
  \multicolumn{1}{c}{.878 ± .0055} &
  \multicolumn{1}{c}{.721 ± .0055} &
  .756 ± .0037 &
  \multicolumn{1}{c}{.0336 ± .00014} &
  \multicolumn{1}{c}{.867 ± .0061} &
  \multicolumn{1}{c}{.713 ± .0053} &
  .751 ± .0042 \\ \cmidrule(l){2-10} 
 &
  Lin Interp &
  \multicolumn{1}{c}{.0246 ± .00009} &
  \multicolumn{1}{c}{.896 ± .0057} &
  \multicolumn{1}{c}{.736 ± .0059} &
  .774 ± .0037 &
  \multicolumn{1}{c}{.0477 ± .00031} &
  \multicolumn{1}{c}{.862 ± .0061} &
  \multicolumn{1}{c}{.713 ± .0051} &
  .752 ± .0042 \\ \cmidrule(l){2-10} 
 &
  FFT &
  \multicolumn{1}{c}{.0490 ± .00015} &
  \multicolumn{1}{c}{.880 ± .0062} &
  \multicolumn{1}{c}{.724 ± .0056} &
  .746 ± .0037 &
  \multicolumn{1}{c}{.0397 ± .00015} &
  \multicolumn{1}{c}{.870 ± .0060} &
  \multicolumn{1}{c}{.702 ± .0055} &
  .752 ± .0037 \\ \cmidrule(l){2-10} 
 &
  BRITS &
  \multicolumn{1}{c}{.0042 ± .00003} &
  \multicolumn{1}{c}{.931 ± .0041} &
  \multicolumn{1}{c}{.748 ± .0050} &
  .795 ± .0036 &
  \multicolumn{1}{c}{.0343 ± .00015} &
  \multicolumn{1}{c}{.871 ± .0055} &
  \multicolumn{1}{c}{.716 ± .0051} &
  .759 ± .0041 \\ \cmidrule(l){2-10} 
 &
  BRITS w/ GAIL &
  \multicolumn{1}{c}{.0378 ± .00015} &
  \multicolumn{1}{c}{.831 ± .0072} &
  \multicolumn{1}{c}{.689 ± .0058} &
  .718 ± .0039 &
  \multicolumn{1}{c}{.5566 ± .00293} &
  \multicolumn{1}{c}{.863 ± .0055} &
  \multicolumn{1}{c}{.691 ± .0056} &
  .738 ± .0039 \\ \cmidrule(l){2-10} 
 &
  NAOMI &
  \multicolumn{1}{c}{.0252 ± .00013} &
  \multicolumn{1}{c}{.897 ± .0051} &
  \multicolumn{1}{c}{.730 ± .0053} &
  .762 ± .0038 &
  \multicolumn{1}{c}{.0493 ± .00025} &
  \multicolumn{1}{c}{.855 ± .0057} &
  \multicolumn{1}{c}{.690 ± .0056} &
  .745 ± .0041 \\ \cmidrule(l){2-10} 
 &
  DeepMVI &
  \multicolumn{1}{c}{.0038 ± .00002} &
  \multicolumn{1}{c}{.929 ± .0039} &
  \multicolumn{1}{c}{.748 ± .0054} &
  .794 ± .0037 &
  \multicolumn{1}{c}{.0316 ± .00012} &
  \multicolumn{1}{c}{.836 ± .0070} &
  \multicolumn{1}{c}{.712 ± .0053} &
  .734 ± .0039 \\ \cmidrule(l){2-10} 
 &
  Van Trans &
  \multicolumn{1}{c}{.0019 ± .00002} &
  \multicolumn{1}{c}{.929 ± .0043} &
  \multicolumn{1}{c}{.747 ± .0053} &
  .792 ± .0037 &
  \multicolumn{1}{c}{.0286 ± .00010} &
  \multicolumn{1}{c}{.848 ± .0068} &
  \multicolumn{1}{c}{.728 ± .0054} &
  .743 ± .0043 \\ \cmidrule(l){2-10} 
 &
  Conv9 Trans &
  \multicolumn{1}{c}{.0026 ± .00002} &
  \multicolumn{1}{c}{.924 ± .0046} &
  \multicolumn{1}{c}{.747 ± .0052} &
  .791 ± .0037 &
  \multicolumn{1}{c}{.0326 ± .00013} &
  \multicolumn{1}{c}{.870 ± .0057} &
  \multicolumn{1}{c}{.715 ± .0052} &
  .752 ± .0041 \\ \cmidrule(l){2-10} 
 &
  BDC Trans &
  \multicolumn{1}{c}{.0017 ± .00001} &
  \multicolumn{1}{c}{.931 ± .0042} &
  \multicolumn{1}{c}{.747 ± .0054} &
  .793 ± .0036 &
  \multicolumn{1}{c}{.0153 ± .00009} &
  \multicolumn{1}{c}{.919 ± .0037} &
  \multicolumn{1}{c}{.735 ± .0050} &
  .773 ± .0039 \\ \midrule
\multirow{14}{*}{30} &
  Mean &
  \multicolumn{1}{c}{.0335 ± .00011} &
  \multicolumn{1}{c}{.847 ± .0060} &
  \multicolumn{1}{c}{.686 ± .0059} &
  .723 ± .0040 &
  \multicolumn{1}{c}{.0340 ± .00014} &
  \multicolumn{1}{c}{.864 ± .0054} &
  \multicolumn{1}{c}{.691 ± .0056} &
  .737 ± .0044 \\ \cmidrule(l){2-10} 
 &
  Lin Interp &
  \multicolumn{1}{c}{.0266 ± .00008} &
  \multicolumn{1}{c}{.869 ± .0051} &
  \multicolumn{1}{c}{.691 ± .0054} &
  .740 ± .0041 &
  \multicolumn{1}{c}{.0487 ± .00032} &
  \multicolumn{1}{c}{.868 ± .0067} &
  \multicolumn{1}{c}{.721 ± .0060} &
  .756 ± .0037 \\ \cmidrule(l){2-10} 
 &
  FFT &
  \multicolumn{1}{c}{.0489 ± .00014} &
  \multicolumn{1}{c}{.840 ± .0065} &
  \multicolumn{1}{c}{.694 ± .0057} &
  .712 ± .0039 &
  \multicolumn{1}{c}{.0384 ± .00015} &
  \multicolumn{1}{c}{.868 ± .0058} &
  \multicolumn{1}{c}{.682 ± .0055} &
  .727 ± .0038 \\ \cmidrule(l){2-10} 
 &
  BRITS &
  \multicolumn{1}{c}{.0051 ± .00003} &
  \multicolumn{1}{c}{.933 ± .0035} &
  \multicolumn{1}{c}{.745 ± .0051} &
  .792 ± .0036 &
  \multicolumn{1}{c}{.0346 ± .00014} &
  \multicolumn{1}{c}{.865 ± .0053} &
  \multicolumn{1}{c}{.697 ± .0054} &
  .748 ± .0041 \\ \cmidrule(l){2-10} 
 &
  BRITS w/ GAIL &
  \multicolumn{1}{c}{.0379 ± .00014} &
  \multicolumn{1}{c}{.767 ± .0077} &
  \multicolumn{1}{c}{.646 ± .0057} &
  .669 ± .0039 &
  \multicolumn{1}{c}{.6916 ± .00279} &
  \multicolumn{1}{c}{.849 ± .0054} &
  \multicolumn{1}{c}{.674 ± .0053} &
  .724 ± .0040 \\ \cmidrule(l){2-10} 
 &
  NAOMI &
  \multicolumn{1}{c}{.0249 ± .00011} &
  \multicolumn{1}{c}{.875 ± .0053} &
  \multicolumn{1}{c}{.705 ± .0055} &
  .741 ± .0039 &
  \multicolumn{1}{c}{.0513 ± .00024} &
  \multicolumn{1}{c}{.841 ± .0061} &
  \multicolumn{1}{c}{.666 ± .0059} &
  .718 ± .0042 \\ \cmidrule(l){2-10} 
 &
  DeepMVI &
  \multicolumn{1}{c}{.0048 ± .00003} &
  \multicolumn{1}{c}{.925 ± .0038} &
  \multicolumn{1}{c}{.743 ± .0053} &
  .790 ± .0036 &
  \multicolumn{1}{c}{.0320 ± .00012} &
  \multicolumn{1}{c}{.834 ± .0072} &
  \multicolumn{1}{c}{.698 ± .0054} &
  .736 ± .0037 \\ \cmidrule(l){2-10} 
 &
  Van Trans &
  \multicolumn{1}{c}{.0027 ± .00002} &
  \multicolumn{1}{c}{.926 ± .0043} &
  \multicolumn{1}{c}{.745 ± .0054} &
  .791 ± .0037 &
  \multicolumn{1}{c}{.0289 ± .00010} &
  \multicolumn{1}{c}{.852 ± .0064} &
  \multicolumn{1}{c}{.704 ± .0057} &
  .735 ± .0041 \\ \cmidrule(l){2-10} 
 &
  Conv9 Trans &
  \multicolumn{1}{c}{.0038 ± .00003} &
  \multicolumn{1}{c}{.917 ± .0045} &
  \multicolumn{1}{c}{.743 ± .0055} &
  .787 ± .0037 &
  \multicolumn{1}{c}{.0329 ± .00013} &
  \multicolumn{1}{c}{.863 ± .0056} &
  \multicolumn{1}{c}{.691 ± .0055} &
  .741 ± .0043 \\ \cmidrule(l){2-10} 
 &
  BDC Trans &
  \multicolumn{1}{c}{.0021 ± .00002} &
  \multicolumn{1}{c}{.930 ± .0041} &
  \multicolumn{1}{c}{.745 ± .0053} &
  .792 ± .0037 &
  \multicolumn{1}{c}{.0174 ± .00009} &
  \multicolumn{1}{c}{.917 ± .0038} &
  \multicolumn{1}{c}{.719 ± .0053} &
  .760 ± .0039 \\ \midrule
\multirow{14}{*}{40} &
  Mean &
  \multicolumn{1}{c}{.0335 ± .00011} &
  \multicolumn{1}{c}{.806 ± .0074} &
  \multicolumn{1}{c}{.646 ± .0060} &
  .683 ± .0042 &
  \multicolumn{1}{c}{.0345 ± .00014} &
  \multicolumn{1}{c}{.851 ± .0054} &
  \multicolumn{1}{c}{.684 ± .0058} &
  .736 ± .0041 \\ \cmidrule(l){2-10} 
 &
  Lin Interp &
  \multicolumn{1}{c}{.0287 ± .00008} &
  \multicolumn{1}{c}{.850 ± .0057} &
  \multicolumn{1}{c}{.684 ± .0058} &
  .734 ± .0041 &
  \multicolumn{1}{c}{.0496 ± .00031} &
  \multicolumn{1}{c}{.835 ± .0075} &
  \multicolumn{1}{c}{.696 ± .0060} &
  .728 ± .0038 \\ \cmidrule(l){2-10} 
 &
  FFT &
  \multicolumn{1}{c}{.0485 ± .00014} &
  \multicolumn{1}{c}{.796 ± .0068} &
  \multicolumn{1}{c}{.657 ± .0058} &
  .668 ± .0042 &
  \multicolumn{1}{c}{.0372 ± .00014} &
  \multicolumn{1}{c}{.840 ± .0062} &
  \multicolumn{1}{c}{.668 ± .0057} &
  .712 ± .0038 \\ \cmidrule(l){2-10} 
 &
  BRITS &
  \multicolumn{1}{c}{.0063 ± .00003} &
  \multicolumn{1}{c}{.928 ± .0036} &
  \multicolumn{1}{c}{.738 ± .0053} &
  .787 ± .0036 &
  \multicolumn{1}{c}{.0349 ± .00014} &
  \multicolumn{1}{c}{.852 ± .0054} &
  \multicolumn{1}{c}{.693 ± .0056} &
  .742 ± .0041 \\ \cmidrule(l){2-10} 
 &
  BRITS w/ GAIL &
  \multicolumn{1}{c}{.0380 ± .00014} &
  \multicolumn{1}{c}{.699 ± .0081} &
  \multicolumn{1}{c}{.600 ± .0057} &
  .619 ± .0041 &
  \multicolumn{1}{c}{.8261 ± .00253} &
  \multicolumn{1}{c}{.836 ± .0061} &
  \multicolumn{1}{c}{.661 ± .0062} &
  .711 ± .0040 \\ \cmidrule(l){2-10} 
 &
  NAOMI &
  \multicolumn{1}{c}{.0251 ± .00010} &
  \multicolumn{1}{c}{.842 ± .0068} &
  \multicolumn{1}{c}{.676 ± .0056} &
  .716 ± .0040 &
  \multicolumn{1}{c}{.0514 ± .00022} &
  \multicolumn{1}{c}{.797 ± .0085} &
  \multicolumn{1}{c}{.653 ± .0060} &
  .693 ± .0042 \\ \cmidrule(l){2-10} 
 &
  DeepMVI &
  \multicolumn{1}{c}{.0064 ± .00003} &
  \multicolumn{1}{c}{.914 ± .0043} &
  \multicolumn{1}{c}{.738 ± .0052} &
  .784 ± .0036 &
  \multicolumn{1}{c}{.0325 ± .00012} &
  \multicolumn{1}{c}{.821 ± .0067} &
  \multicolumn{1}{c}{.691 ± .0056} &
  .731 ± .0039 \\ \cmidrule(l){2-10} 
 &
  Van Trans &
  \multicolumn{1}{c}{.0041 ± .00002} &
  \multicolumn{1}{c}{.921 ± .0043} &
  \multicolumn{1}{c}{.743 ± .0050} &
  .788 ± .0037 &
  \multicolumn{1}{c}{.0302 ± .00010} &
  \multicolumn{1}{c}{.839 ± .0062} &
  \multicolumn{1}{c}{.689 ± .0057} &
  .719 ± .0045 \\ \cmidrule(l){2-10} 
 &
  Conv9 Trans &
  \multicolumn{1}{c}{.0058 ± .00003} &
  \multicolumn{1}{c}{.902 ± .0050} &
  \multicolumn{1}{c}{.738 ± .0052} &
  .782 ± .0037 &
  \multicolumn{1}{c}{.0332 ± .00013} &
  \multicolumn{1}{c}{.851 ± .0055} &
  \multicolumn{1}{c}{.684 ± .0058} &
  .738 ± .0041 \\ \cmidrule(l){2-10} 
 &
  BDC Trans &
  \multicolumn{1}{c}{.0030 ± .00002} &
  \multicolumn{1}{c}{.927 ± .0042} &
  \multicolumn{1}{c}{.742 ± .0052} &
  .789 ± .0037 &
  \multicolumn{1}{c}{.0203 ± .00009} &
  \multicolumn{1}{c}{.907 ± .0039} &
  \multicolumn{1}{c}{.705 ± .0053} &
  .746 ± .0040 \\ \midrule
\multirow{14}{*}{50} &
  Mean &
  \multicolumn{1}{c}{.0335 ± .00011} &
  \multicolumn{1}{c}{.753 ± .0075} &
  \multicolumn{1}{c}{.611 ± .0059} &
  .638 ± .0043 &
  \multicolumn{1}{c}{.0351 ± .00014} &
  \multicolumn{1}{c}{.847 ± .0061} &
  \multicolumn{1}{c}{.681 ± .0053} &
  .728 ± .0042 \\ \cmidrule(l){2-10} 
 &
  Lin Interp &
  \multicolumn{1}{c}{.0311 ± .00008} &
  \multicolumn{1}{c}{.849 ± .0058} &
  \multicolumn{1}{c}{.680 ± .0054} &
  .731 ± .0041 &
  \multicolumn{1}{c}{.0506 ± .00031} &
  \multicolumn{1}{c}{.790 ± .0076} &
  \multicolumn{1}{c}{.666 ± .0060} &
  .698 ± .0039 \\ \cmidrule(l){2-10} 
 &
  FFT &
  \multicolumn{1}{c}{.0476 ± .00014} &
  \multicolumn{1}{c}{.734 ± .0076} &
  \multicolumn{1}{c}{.619 ± .0062} &
  .623 ± .0043 &
  \multicolumn{1}{c}{.0364 ± .00013} &
  \multicolumn{1}{c}{.834 ± .0062} &
  \multicolumn{1}{c}{.662 ± .0054} &
  .698 ± .0042 \\ \cmidrule(l){2-10} 
 &
  BRITS &
  \multicolumn{1}{c}{.0080 ± .00003} &
  \multicolumn{1}{c}{.915 ± .0040} &
  \multicolumn{1}{c}{.731 ± .0053} &
  .779 ± .0035 &
  \multicolumn{1}{c}{.0350 ± .00013} &
  \multicolumn{1}{c}{.851 ± .0057} &
  \multicolumn{1}{c}{.692 ± .0053} &
  .736 ± .0044 \\ \cmidrule(l){2-10} 
 &
  BRITS w/ GAIL &
  \multicolumn{1}{c}{.0380 ± .00014} &
  \multicolumn{1}{c}{.620 ± .0083} &
  \multicolumn{1}{c}{.563 ± .0060} &
  .578 ± .0044 &
  \multicolumn{1}{c}{.9647 ± .00183} &
  \multicolumn{1}{c}{.825 ± .0059} &
  \multicolumn{1}{c}{.661 ± .0057} &
  .711 ± .0040 \\ \cmidrule(l){2-10} 
 &
  NAOMI &
  \multicolumn{1}{c}{.0261 ± .00010} &
  \multicolumn{1}{c}{.809 ± .0068} &
  \multicolumn{1}{c}{.642 ± .0053} &
  .691 ± .0038 &
  \multicolumn{1}{c}{.0519 ± .00020} &
  \multicolumn{1}{c}{.775 ± .0073} &
  \multicolumn{1}{c}{.638 ± .0057} &
  .679 ± .0043 \\ \cmidrule(l){2-10} 
 &
  DeepMVI &
  \multicolumn{1}{c}{.0090 ± .00004} &
  \multicolumn{1}{c}{.891 ± .0052} &
  \multicolumn{1}{c}{.726 ± .0052} &
  .772 ± .0036 &
  \multicolumn{1}{c}{.0329 ± .00012} &
  \multicolumn{1}{c}{.820 ± .0070} &
  \multicolumn{1}{c}{.686 ± .0051} &
  .725 ± .0041 \\ \cmidrule(l){2-10} 
 &
  Van Trans &
  \multicolumn{1}{c}{.0065 ± .00003} &
  \multicolumn{1}{c}{.907 ± .0044} &
  \multicolumn{1}{c}{.733 ± .0052} &
  .780 ± .0037 &
  \multicolumn{1}{c}{.0318 ± .00011} &
  \multicolumn{1}{c}{.830 ± .0067} &
  \multicolumn{1}{c}{.680 ± .0054} &
  .705 ± .0044 \\ \cmidrule(l){2-10} 
 &
  Conv9 Trans &
  \multicolumn{1}{c}{.0088 ± .00004} &
  \multicolumn{1}{c}{.881 ± .0058} &
  \multicolumn{1}{c}{.727 ± .0056} &
  .770 ± .0037 &
  \multicolumn{1}{c}{.0334 ± .00012} &
  \multicolumn{1}{c}{.847 ± .0061} &
  \multicolumn{1}{c}{.682 ± .0055} &
  .729 ± .0042 \\ \cmidrule(l){2-10} 
 &
  BDC Trans &
  \multicolumn{1}{c}{.0054 ± .00003} &
  \multicolumn{1}{c}{.916 ± .0042} &
  \multicolumn{1}{c}{.736 ± .0051} &
  .782 ± .0038 &
  \multicolumn{1}{c}{.0235 ± .00010} &
  \multicolumn{1}{c}{.888 ± .0049} &
  \multicolumn{1}{c}{.691 ± .0055} &
  .728 ± .0040 \\ \bottomrule
\end{tabular}%
}
\caption{For the sake of completeness, these are the full tabulated results from the Union of Leads PTB-XL ECG dataset for the Cardiac Classification task with Macro AUC values with 95\% CI. Assume all other results, besides this table, presented with ECG cardiac classification task are with the Lead I only PTB-XL dataset.}
\label{tab:ptbxlunion}
\end{table}

\FloatBarrier
\clearpage 
\section{Dataset and License details}
The code repository for our benchmarking challenge can be found here: \textcolor{cyan}{\url{www.github.com/rehg-lab/pulseimpute}}   and licensed under the MIT License. The intended use of our curated datasets and missingness patterns is to be used in conjunction with our PulseImpute challenge framework. The training and evaluation procedures for our challenge can be found in our code repo with descriptions in the main text and Appendix A4.

Our curated datasets and missingness patterns can be found linked \textcolor{cyan}{\url{www.doi.org/10.5281/zenodo.7129965}}, and we license them under the Creative Commons Attribution 4.0 International. The ECG and PPG waveforms are a form of personal data, but the identifiers have been removed and its public redistribution is in public interest. The data we provide also does not contain any offensive content. We, the authors, bear all responsibility to withdraw our paper and data in case of violation of licensing or patient privacy rights, and confirmation of the data license. The curated data and missingness patterns are organized as shown below: \\

\dirtree{%
.1 /pulseimpute\_data/.
.2 README.md.
.2 missingness\_patterns/.
.3 mHealth\_missing\_ecg/.
.4 missing\_ecg\_train.csv.
.4 missing\_ecg\_val.csv.
.4 missing\_ecg\_test.csv.
.3 mHealth\_missing\_ppg/.
.4 missing\_ppg\_train.csv.
.4 missing\_ppg\_val.csv.
.4 missing\_ppg\_test.csv.
.2 waveforms/.
.3 mimic\_ecg/.
.4 mimic\_ecg\_train.npy.
.4 mimic\_ecg\_val.npy.
.4 mimic\_ecg\_test.npy.
.4 MIMIC\_III\_ECG\_filenames.txt.
.3 mimic\_ppg/.
.4 mimic\_ppg\_train.npy.
.4 mimic\_ppg\_val.npy.
.4 mimic\_ppg\_test.npy.
.4 MIMIC\_III\_PPG\_filenames.txt.
.3 ptbxl\_ecg/.
.4 scp\_statements.csv.
.4 ptbxl\_database.csv.
.4 ptbxl\_ecg.npy.
} 

The data is all stored as .npy files, with each row corresponding to a 100 Hz waveform. The missingness patterns are stored in csv files, with each row as a list of tuples of size 2, which represent the binary missingness pattern time-series. The first item in the tuple corresponds to missing (0) or not missing (1) with the second entry corresponding to the length of samples (in 100 Hz) that the missing or not missingness segment lasts.  Each of MIMIC-III curated data and missingness patterns have been split into 80/10/10 training/validation/testing splits accordingly.  If we concatenate train, validation, and test .npy files in that order, each data index corresponds to the file name at the corresponding line in the MIMIC\_III\_ECG\_filenames.txt or MIMIC\_III\_PPG\_filenames.txt file.

For the cardiac classification tasks on the PTB-XL data, the labels can be found in the ptbxl\_database.csv file and the waveform data in the ptbxl\_ecg.npy.  We use the original paper's  splits to divide into 40/10/50 training/validation/testing splits. Imputation models and downstream cardiac classification models are trained with the 40/10 split, with the classification model training on the clean non-imputed data. Then classification runs inference on the imputed test data in the 50 split to evaluate imputation quality. This large test split was done to allow for future work where the classification model trains directly on imputed data.

Our datasets originate from the curation of two different datasets, MIMIC-III Waveform \cite{moody2020mimicWAVEFORM} and PTB-XL \cite{ptbxl}. MIMIC-III Waveform uses the Open Data Commons Open Database License v1.0 (linked \textcolor{cyan}{\href{https://physionet.org/content/mimic3wdb/view-license/1.0/}{here}}), which explicitly allows for the creation and distribution of derivative databases, which we have done via our curation described in Section \ref{sec:curation}. We have attributed the data to its original source throughout our paper, by citing   \cite{moody2020mimicWAVEFORM}.  PTB-XL uses the Creative Commons Attribution 4.0 International Public License (linked \textcolor{cyan}{\href{https://physionet.org/content/ptb-xl/view-license/1.0.1/}{here}}), which explicitly allows for adaptation and redistribution of the data, and we have attributed the data to its original source throughout our paper, by citing \cite{ptbxl}. The missingness patterns for PPG were extracted from analyzing PPG-DaLiA, which is hosted on the UCI Machine Learning Repository (linked \textcolor{cyan}{\href{https://archive.ics.uci.edu/ml/datasets/PPG-DaLiA}{here}}), and does not have an explicit license, but states others may use the dataset for scientific, non-commercial purposes, provided that credit is given, which we have done throughout our paper, by citing \cite{reiss2019ppgdalia}. The binary missingness patterns for ECG were extracted from \cite{chatterjee2020smokingopp}, our mHealth study. For additional documentation for the datasets we have mentioned, please see the original publications that the datasets originate from \cite{moody2020mimicWAVEFORM, ptbxl, reiss2019ppgdalia, chatterjee2020smokingopp}.

\newpage

\phantomsection 
\addcontentsline{toc}{section}{Appendix References}
\renewcommand\refname{Appendix References}
\printbibliography

\end{refsection}

\end{document}